\documentclass[10pt,twocolumn,letterpaper]{article}

\usepackage{cvpr}
\usepackage{times}
\usepackage{epsfig}
\usepackage{graphicx}
\usepackage{amsmath}
\usepackage{amssymb}

\usepackage{bm, url}
\usepackage[usenames]{color}
\usepackage{icomma}
\usepackage[ruled,lined]{algorithm2e}

\newcommand{\itm}{a}
\renewcommand{\i}{{\itm}}
\newcommand{\supp}{{\mathrm{supp}}}
\newcommand{\conf}{{\mathrm{conf}}}

\cvprfinalcopy %

\def\cvprPaperID{****} %

\ifcvprfinal\fi

\begin{document}

\title{Mid-level Deep Pattern Mining\thanks{The
    first two authors contributed to this work equally.
    This work was in part funded by
the Data to Decisions Cooperative Research Centre, Australia.
Correspondence should be addressed to C. Shen (e-mail: {\tt chhshen@gmail.com}).
}
}

\author{
  Yao Li$ ^{1,3}$
  \hspace{.51cm}
  Lingqiao Liu$ ^{1}$
  \hspace{.51cm}
    Chunhua Shen$ ^{1,2} $
    \hspace{.51cm}
  Anton van den Hengel$^{1,2} $
 \\
$ ^1 $The University of Adelaide \hspace{.51cm} %
 $ ^2 $Australian Centre for Robotic Vision
 \hspace{.51cm}
 $ ^3 $NICTA
}

\maketitle
\thispagestyle{empty}

\begin{abstract}
Mid-level visual element discovery aims to find clusters of image patches that are both representative and discriminative.
In this work, we study this problem from the prospective of pattern mining  while relying on the recently popularized Convolutional Neural Networks (CNNs).
Specifically, we find that for an image patch, activations extracted from the first fully-connected layer of CNNs have two appealing properties which enable its seamless integration with pattern mining.
Patterns are then discovered from a large number of CNN activations of image patches through the well-known association rule mining.
When we retrieve and visualize image patches with the same pattern (See Fig.~\ref{fig:Name_that_object}), surprisingly, they are not only visually similar but also semantically consistent.
We apply our approach to scene and object classification tasks, and demonstrate that our approach outperforms all previous works on mid-level visual element discovery by a sizeable margin with far fewer elements being used.
Our approach also outperforms or matches recent works using CNN for these tasks.
Source code of the complete system is available online.
\end{abstract}

\section{Introduction}

\begin{figure}[t]
\vspace{-0.0cm}
\begin{center}
\scalebox{0.9}{
\begin{tabular}{@{}c@{}c@{}c@{}c@{}c@{}c@{}c}
\includegraphics[width=0.5\linewidth]{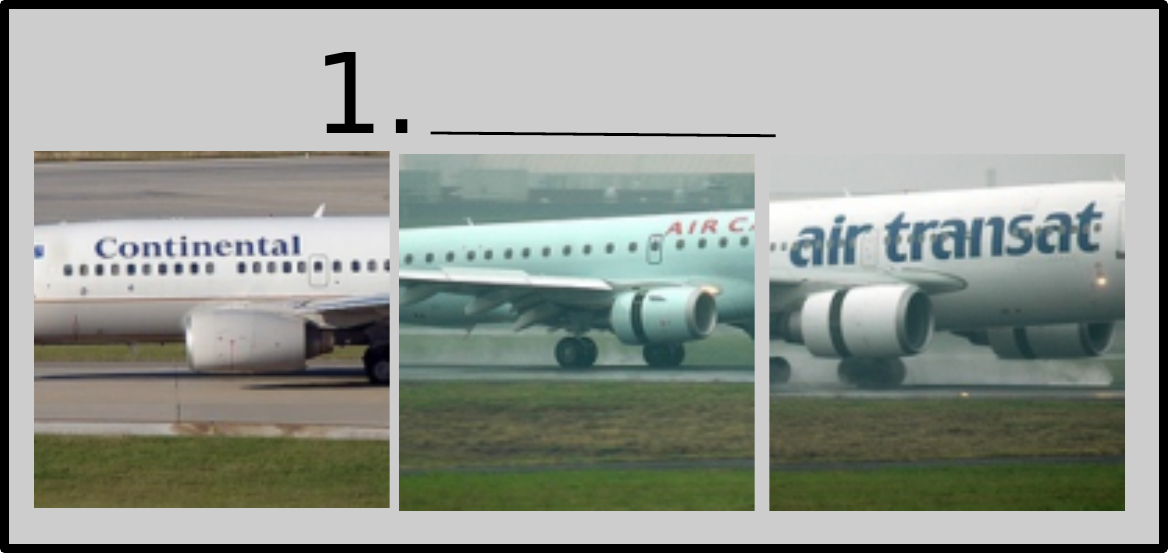} \ &
\includegraphics[width=0.5\linewidth]{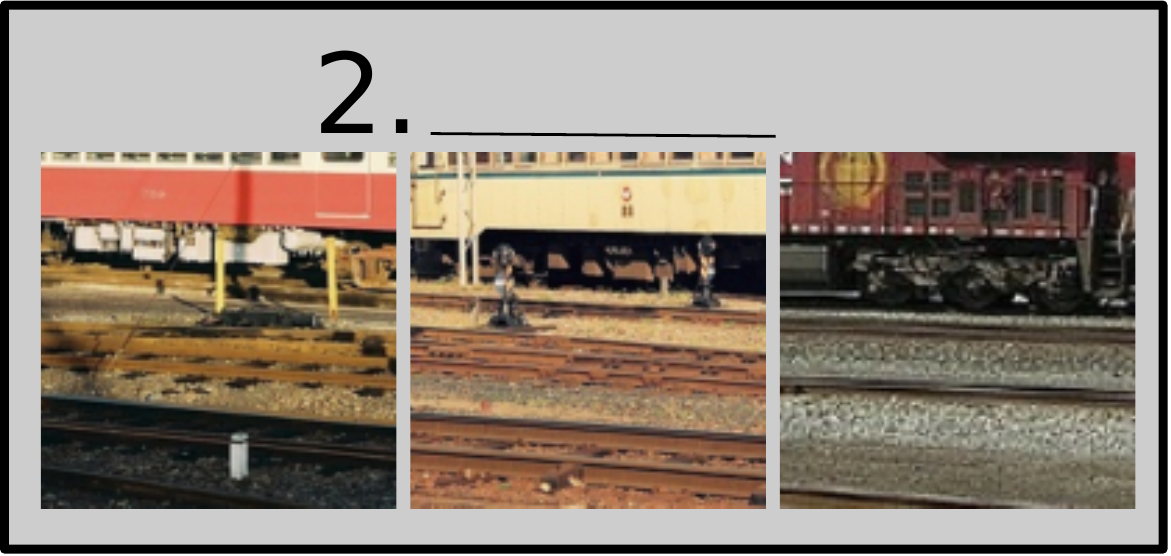} \\
\includegraphics[width=0.5\linewidth]{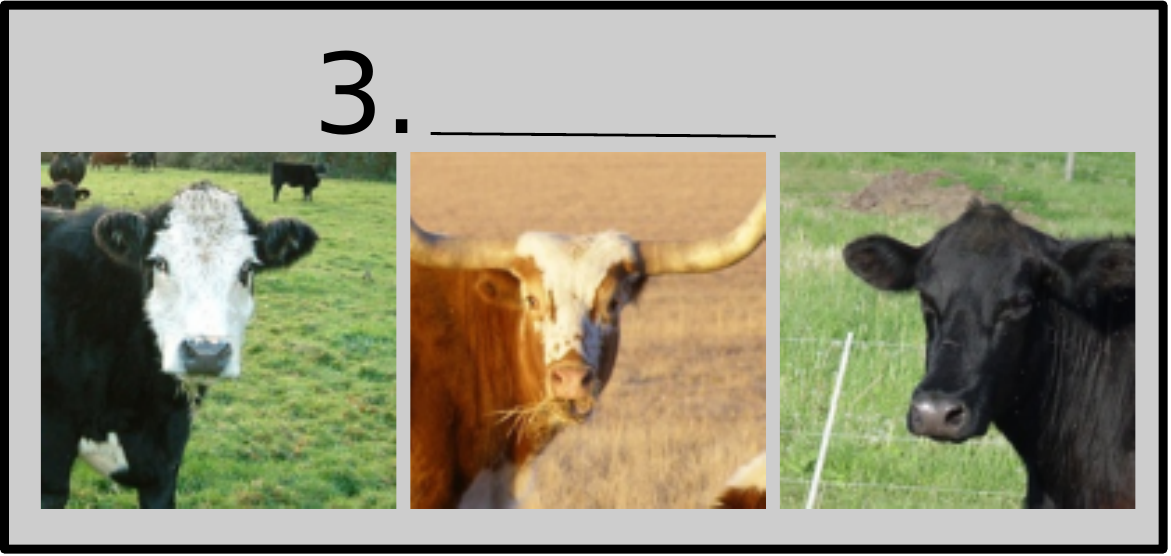} \ &
\includegraphics[width=0.5\linewidth]{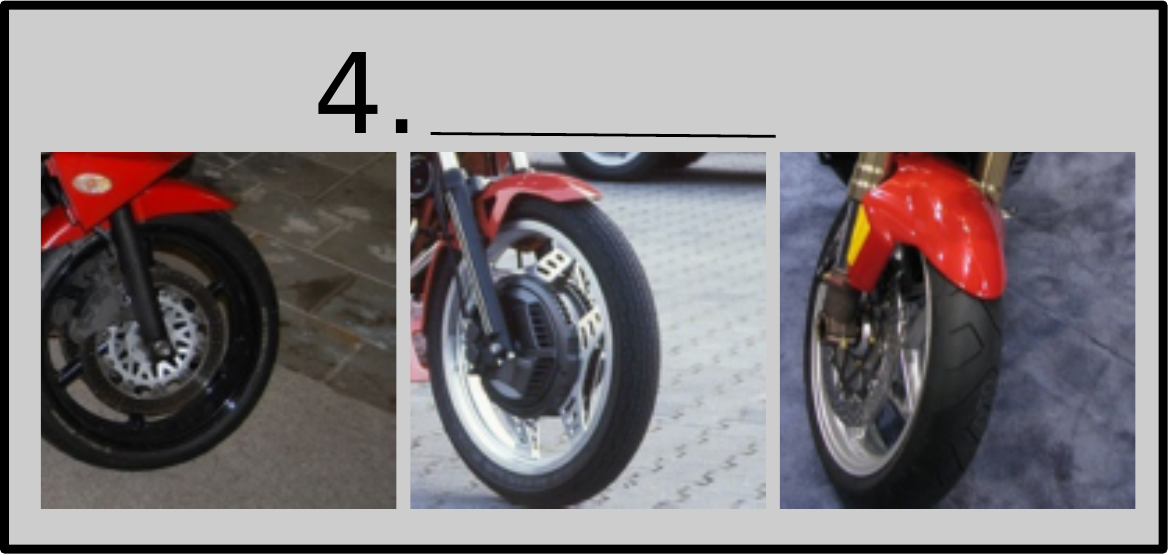} \\
\includegraphics[width=0.5\linewidth]{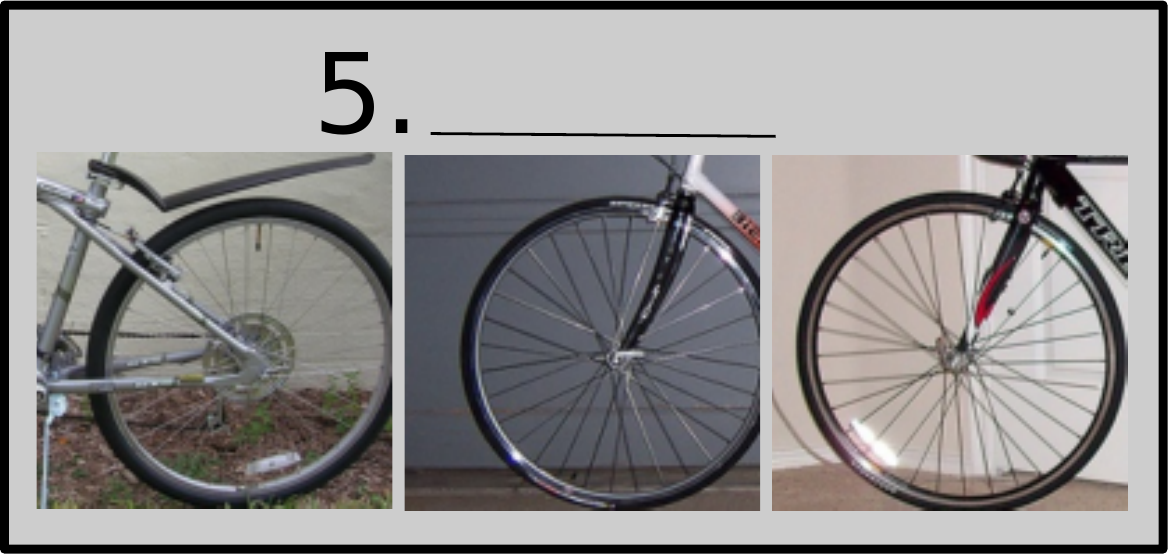} \ &
\includegraphics[width=0.5\linewidth]{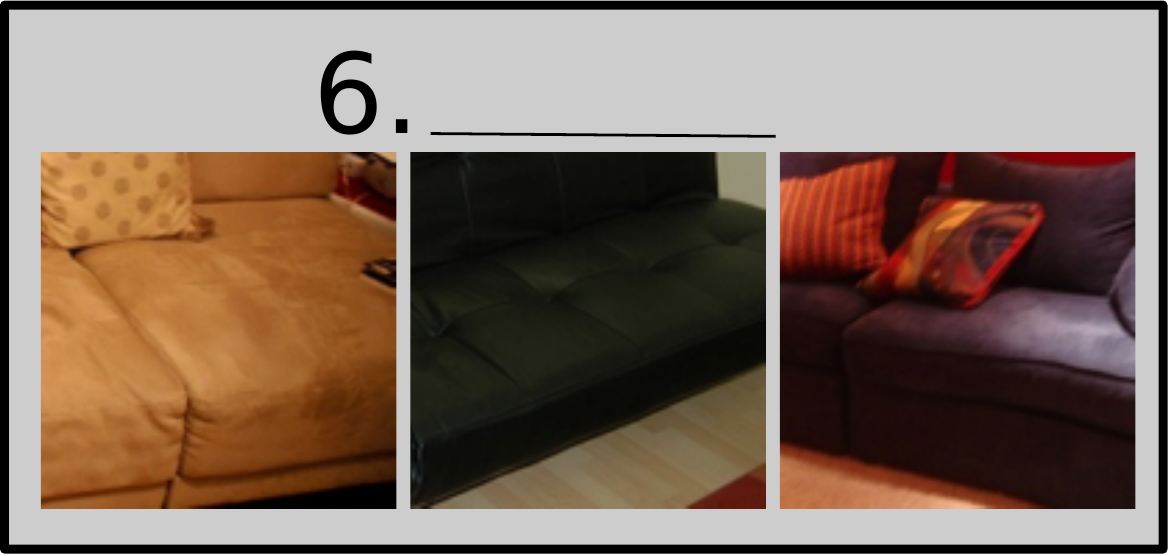} \\
\end{tabular}
}
\end{center}
\caption{\emph{Name that Object:} Given some mid-level visual elements discovered by our algorithm from the PASCAL VOC 2007 dataset, can you guess what categories are they from? (answer key below)}
\label{fig:Name_that_object} \vspace{-0.39cm}
\end{figure}

Mid-level visual elements, which are clusters of image patches rich in semantic meaning, were proposed by Singh~\etal~\cite{DBLP:conf/eccv/SinghGE12} with the aim of replacing low-level visual words (play the game in Fig.~\ref{fig:Name_that_object} and then check your answers below\footnote{Answer key: 1.aeroplane, 2.train, 3.cow, 4.motorbike, 5.bike, 6.sofa.}).
In this pioneering work, mid-level visual elements must meet two requirements, that is, representativeness and discriminativeness.
Representativeness means mid-level visual elements should frequently occur in the target category, while discriminativeness implies that they should be visually discriminative against the natural world.
The discovery of mid-level visual elements has boosted performance in a variety of vision tasks, including image classification~\cite{DBLP:conf/eccv/SinghGE12,DBLP:conf/nips/DoerschGE13,
DBLP:conf/cvpr/JunejaVJZ13,
DBLP:conf/iccv/SunP13,DBLP:conf/cvpr/LiWT13,DBLP:conf/icml/WangWBLT13, DBLP:conf/eccv/BossardGG14},
action recognition~\cite{DBLP:conf/cvpr/JainGRD13,DBLP:conf/cvpr/WangQT13},
discovering stylistic elements~\cite{DBLP:journals/tog/DoerschSGSE12,DBLP:conf/iccv/LeeEM13}, geometry estimation~\cite{DBLP:conf/iccv/FouheyGH13}
and 2D-3D alignment~\cite{DBLP:journals/tog/AubryRS14,DBLP:conf/cvpr/AubryMERSJ14}.

Originally motivated by market basket analysis, association rule mining is a well-known pattern mining algorithm that aims to discover a collection of if-then rules (\ie, association rules) from a large number of records named transactions.
The main advantage of association rule mining lies in its ability to process ``Big Data'': association rules can be discovered from millions of transactions efficiently.
In the context of mid-level visual element discovery, as noted by Doersch~\etal~\cite{DBLP:conf/nips/DoerschGE13}, finding discriminative patches usually involves searching through tens of thousands of patches, which is a bottleneck in recent works.
In this sense, if appropriately used, association rule mining can be an appealing solution for handling ``Big Data'' in mid-level visual element discovery.

In this paper, building on the well-known association rule mining, we propose a pattern mining algorithm, \emph{Mid-level Deep Pattern Mining} (MDPM), to study the problem of mid-level visual element discovery.
This approach is particularly appealing because the specific properties of activation extracted from the fully-connected layer of a Convolutional Neural Network (CNN) allow them to be seamlessly
integrated with association rule mining, which enables the discovery of category-specific patterns from a large number of image patches .
Moreover, we find that two requirements of mid-level visual elements, representativeness and discriminativeness, can be easily fulfilled by association rule mining.
When we visualize image patches with the same pattern (mid-level visual element in our scenario), it turns out that they are not only visually similar, but also semantically consistent (see Fig.~\ref{fig:Name_that_object}).

To our knowledge, hand-crafted features, typically HOG~\cite{DBLP:conf/cvpr/DalalT05}, are used as feature descriptors for image patches in all current methods of mid-level visual element discovery.
Vondrick~\etal~\cite{DBLP:conf/iccv/VondrickKMT13},
however, have illustrated the limitations of HOG, implying that
HOG may be too lossy a descriptor to achieve high recognition performance.
In this sense, an extra bonus of our formulation lies in that we are now relying on CNN activations, a more appealing alternative than the hand-crafted HOG, as indicated in recent works~\cite{DBLP:conf/nips/KrizhevskySH12,6910029,
DBLP:conf/cvpr/GirshickDDM14,DBLP:conf/cvpr/OquabBL14,
DBLP:conf/eccv/Agrawal14,DBLP:conf/eccv/GongWGL14}.

One issue must be considered before using any pattern mining algorithms, that is, they have two strict requirements for the transactions that they can process (Sec.~\ref{subsec:FIM_revisited}).
Thanks to two appealing properties of CNN activations (Sec.~\ref{sec:two_properties}), these two requirements are effortlessly fulfilled in the proposed MDPM algorithm (Sec.~\ref{sec:DFIM}).

To show the effectiveness of the proposed MDPM algorithm, we apply it to scene and generic object classification tasks (Sec.~\ref{sec:application}).
Specifically, after retrieving visual elements from the discovered patterns, we train element detectors and generate new feature representations using these detectors.
We demonstrate that we achieve classification results which not only outperform all current methods in mid-level visual element discovery by a noticeable margin with far fewer elements used, but also outperform or match the performance of state-of-the-art using CNNs for the same task.

In summary, our contributions are twofold.
\begin{enumerate}
\itemsep -1mm
\item We formulate mid-level visual element discovery from the prospective of pattern mining, finding that its two requirements,
representativeness and discriminativeness, can be easily fulfilled by the well-known association rule mining algorithm.
\item We present two properties of CNN activations that allow seamless integration with association rule mining, avoiding the limitations of pattern mining algorithms.
\end{enumerate}

The source code of the complete system is available at \url{http://goo.gl/u5q8ZX}.

\vspace{2mm}
\noindent To extract CNN activations, we rely on the publicly available \texttt{caffe}~\cite{Jia13caffe} reference model which was pre-trained on the ImageNet~\cite{DBLP:conf/cvpr/DengDSLL009}. More specially, given a mean-subtracted patch or image, we
resize it to the size of $227\times227$ and pass it to the \texttt{caffe} CNN. We extract the non-negative $4096$-dimensional CNN activations from the sixth layer \texttt{fc}$6$ (the first fully-connected layer) after the rectified linear unit (ReLU) transformation.

\section{Related Work}
\label{sec:related_work}
\noindent
\textbf{Mid-level visual element discovery.}
Mid-level visual element discovery aims to discover clusters of image patches that are both representative and discriminative.
Recent studies on this topic have shown that mid-level visual elements can be used for image classification~\cite{DBLP:conf/eccv/SinghGE12,DBLP:conf/nips/DoerschGE13,
DBLP:conf/cvpr/JunejaVJZ13,
DBLP:conf/iccv/SunP13,DBLP:conf/cvpr/LiWT13,DBLP:conf/icml/WangWBLT13, DBLP:conf/eccv/BossardGG14}.
The process typically proceeds as follows. Firstly, mining visual elements and training element detectors. Secondly, generate new feature
representations using these element detectors.
Various methods have been proposed for the first step, such as cross validation training of element detectors~\cite{DBLP:conf/eccv/SinghGE12}, ranking and selecting exemplar detectors on the validation set~\cite{DBLP:conf/cvpr/JunejaVJZ13} and discriminative mode seeking~\cite{DBLP:conf/nips/DoerschGE13}.
\vspace{0.2cm}
\newline
\textbf{Convolutional Neural Networks.}
Although proposed by LeCun \etal~\cite{DBLP:conf/nips/CunBDHHHJ89} for solving the handwritten digit recognition in the late '80s, CNNs have regained popularity having shown very promising result in the ILSVRC challenge~\cite{DBLP:journals/corr/RussakovskyDSKSMHKKBBF14}.
In the benchmark CNN architecture of Krizhevsky~\etal~\cite{DBLP:conf/nips/KrizhevskySH12}, raw pixels first pass through five convolutional layers where responses of filters are max-pooled in sequence, before producing an activation of $4096$ dimensions at each of the two fully-connected layers.
Recent studies~\cite{6910029,DBLP:conf/icml/DonahueJVHZTD14} have demonstrated that the $4096$-dimensional activation extracted from the fully-connected layer is an excellent representation for general recognition tasks.
\vspace{0.2cm}
\newline
\textbf{Pattern mining in vision.} Pattern mining techniques have been studied primarily amongst the data mining community, but a growing number of applications can be found in vision, such as image classification~\cite{DBLP:conf/iccv/QuackFLG07,
DBLP:conf/cvpr/YuanYW11,DBLP:conf/eccv/FernandoFT12,
DBLP:conf/cvpr/Voravuthikunchai14}, action recognition~\cite{DBLP:conf/cvpr/WangLWY12} and recognizing human-object interaction~\cite{DBLP:conf/cvpr/YaoF10}.
The main advantage of pattern mining lies its ability to process massive volumes of data, which is particularly appealing in this era of information explosion.
\begin{figure*}[t]
\vspace{-0.0cm}
\begin{center}
\scalebox{0.9}{
\begin{tabular}{@{}c}
\includegraphics[width=1\linewidth]{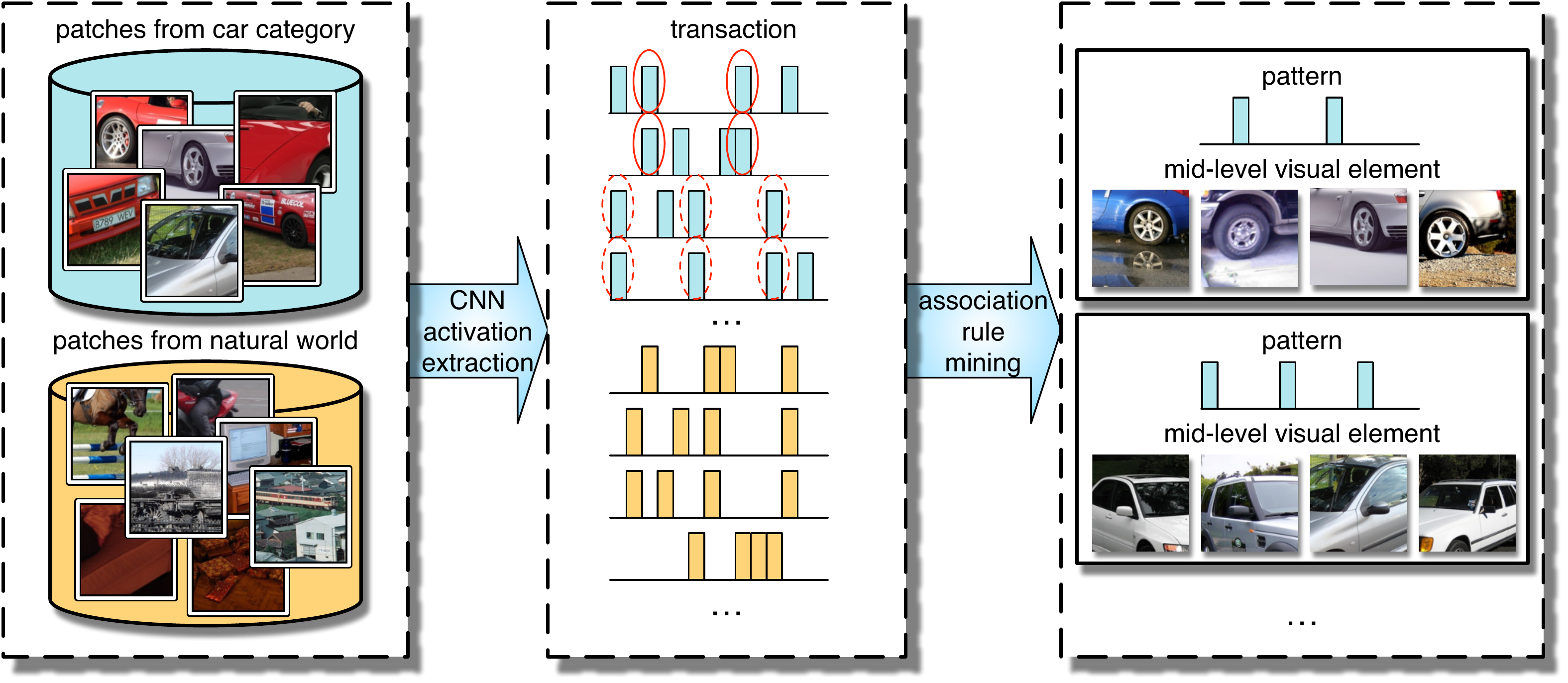} \\
\end{tabular}
}
\end{center}
\caption{Pipeline of mid-level deep pattern mining. Given image patches sampled from both the target category (\eg car) and the natural world, we represent each as a transaction after extracting their CNN activation. Patterns are then discovered through association rule mining. Mid-level visual elements are discovered by retrieving image patches with the same patterns.}
\label{fig:pipleline} \vspace{-0.39cm}
\end{figure*}

\section{Properties of CNN activations of patches}
\label{sec:two_properties}
In this section we provide a detailed analysis of the performance of CNN activations on the MIT Indoor dataset~\cite{DBLP:conf/cvpr/QuattoniT09},
from which we are able to conclude two important properties thereof.
These two properties are key ingredients in the proposed MDPM algorithm in Sec.~\ref{sec:DFIM}.

We first sample $128\times128$ patches with a stride of $32$ pixels from each image. Then, for each image patch, we extract its $4096$-dimensional CNN activation using \texttt{caffe}.
To generate final the feature representation for an image, we consider two methods as follows.
\begin{enumerate}
\itemsep -1mm
\item \textbf{CNN-Sparsified.} For each $4096$-dimensional CNN activation of an image patch, keep only the $k$ largest
dimensions (in terms of magnitude) and set the remaining elements
to zero.
The final feature representation for an image is the outcome of max pooling on the revised CNN activations.
\item \textbf{CNN-Binarized.} For each $4096$-dimensional CNN activation of an image patch, set its $k$ largest
dimensions to one and the remaining elements to zero.
The final feature representation for an image is the outcome of max pooling on these binarized CNN activations.
\end{enumerate}
For each of the above cases, we train a multi-class linear SVM classifier in the one-vs-all fashion.
The classification accuracy achieved by each of the above strategies for a range of $k$ values is summarized in Table~\ref{tab:top_k}.
In comparison, our baseline feature, which is the outcome of max pooling on CNN activations of all patches in an image, gives an accuracy of $65.15\%$.
Analysing the results in Table~\ref{tab:top_k} leads to 2 conclusions:
\begin{table}[t]
\begin{center}
\begin{tabular}{|l|c|c|c|c|}
\hline
 $k$ & $10$ & $20$ & $50$ & $100$\\
\hline
CNN-Sparsified & $50.10$ & $56.33$ & $60.34$ & $61.68$ \\
\hline
CNN-Binarized & $54.34$ & $59.15$ & $61.35$ & $61.29$ \\
\hline
\end{tabular}
\end{center}
\caption{The classification accuracies achieved by the two proposed strategies for keeping the $k$ largest magnitudes of CNN activations of image patches on the MIT Indoor dataset.
}
\label{tab:top_k}
\end{table}
\begin{enumerate}
\itemsep -1mm
\item \textbf{Sparse.} Comparing the performance of ``CNN-Sparsified'' with that of the baseline feature, it is clear that accuracy is reasonably high when using sparsified CNN activations with only $k$ non-zero magnitudes out of $4096$.
\item \textbf{Binary.} Comparing ``CNN-Binarized'' case with its ``CNN-Sparsified'' counterpart, it is observed that CNN activation does not suffer from binarization when $k$ is small, accuracy even increases slightly instead.
\end{enumerate}
\noindent
\textbf{Conclusion.} The above two properties imply that for an image patch, the discriminative information within its CNN activation is mostly embedded in
\emph{the dimension indices of the $k$ largest magnitudes}.

\section{Mid-level deep pattern mining}
\label{sec:DFIM}
In this section, we give details of the proposed MDPM
algorithm, an overview of which is provided in~Fig.~\ref{fig:pipleline}.
We start by introducing some important concepts and terminology in pattern mining.
\begin{figure*}
\begin{center}
\begin{tabular}{@{}c}
\includegraphics[width=1\linewidth]{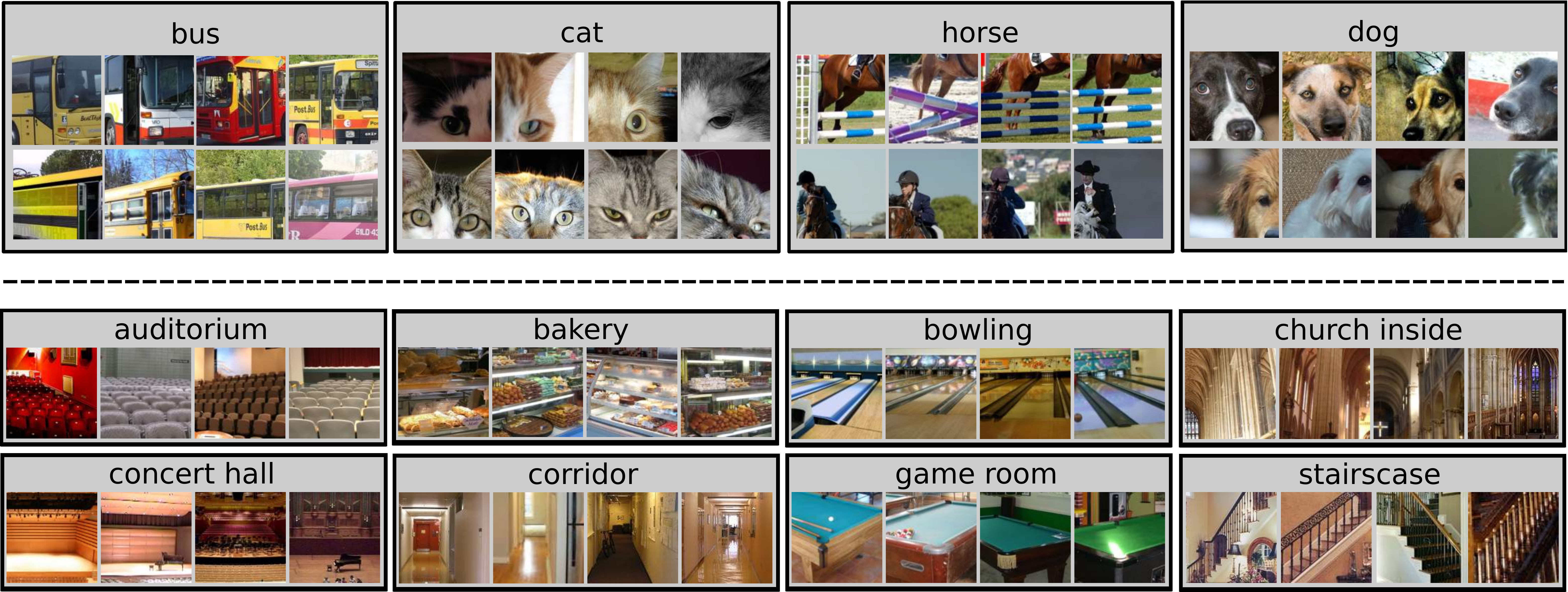} \\
\end{tabular}
\end{center}
\caption{Mid-level visual elements from PASCAL VOC 2007 (top, two per category) and MIT Indoor (bottom, one per category) datasets.}
\label{fig:Name_that_scene}
\end{figure*}
\subsection{Pattern mining revisited}
\label{subsec:FIM_revisited}
\noindent
\textbf{Frequent itemset.}
Let $A=\{\itm_1,\itm_2,\ldots,\itm_K\}$ denote a set of $K$ items.
A transaction $T$ is a subset of $A$, \ie, $T \subseteq A$.
We also define a transaction database $\mathcal{D}=\{T_1,T_2,\ldots,T_N\}$ containing $N$ transactions ($N$ is usually very large).
Given $I$, a subset of $A$, we are interested in the fraction of transactions $T \in \mathcal{D}$ which contain $I$.
The \emph{support} value of $I$ reflects this quantity:
\begin{equation}
\supp(I) = \frac{|\{T|T\in \mathcal{D},I\subseteq T\}|}{N} \in [0,1],
\end{equation}
where $|\cdot|$ measures cardinality.
$I$ is called a \emph{frequent itemset} when $\supp(I)$ is larger than a predefined threshold.\vspace{0.2cm}
\newline
\textbf{Association rule.} An \emph{association rule} $I\rightarrow \i$ implies a relationship between $I$ and an item $\i$.
We are interested in how likely it is that $\i$ is present in the transactions which contain $I$ within $\mathcal{D}$ .
In a typical pattern mining application this might be taken to imply that customers who bought items in $I$ are also likely to buy item $\i$, for instance.
The \emph{confidence} of an association rule $\conf(I\rightarrow \i)$ can be taken to reflect this probability:
\begin{equation}
\label{eq:confidence_definition}
\begin{split}
\conf(I\rightarrow \i) &= \frac{\supp(I\cup\{\i\})}{\supp(I)}\\
    & = \frac{|\{T|T\in \mathcal{D},(I\cup\{\i\})\subseteq T\}|}{|\{T|T\in \mathcal{D},I\subseteq T\}|} \hspace{-0.5mm} \in \hspace{-0.5mm} [0,1].
\end{split}
\end{equation}
In practice, we are interested in ``good'' rules, that is, the confidence of these rules should be reasonably high.\vspace{0.2cm}
\newline
\textbf{Two strict requirements of pattern mining.}
As noted in~\cite{DBLP:conf/cvpr/Voravuthikunchai14}, there are two strict requirements that must be met if we use pattern mining algorithms.
\begin{enumerate}
\itemsep -1mm
\item
Each transaction can only have a small number of items, as the potential search space grows exponentially with the number of items in each transaction.
\item
What is recorded in a transaction is a set of integers, as opposed to the real-valued elements of most vision features (such as SIFT and HOG for example).
\end{enumerate}
\subsection{Transaction creation}
\label{subsec:transaction}
Transactions must be created before any pattern mining algorithms can process.
In our work, as we aim to discover patterns from image patches through pattern mining, \emph{an image patch is utilized to create one transaction}.

The most critical question now is how to transform an image patch into a transaction while maintaining most of its discriminative information.
In this work, we rely on its CNN activation which has two appealing properties (Sec.~\ref{sec:two_properties}).
More specifically, we treat \emph{each dimension index of CNN activation} as an item ($4096$ items in total).
Thanks to two properties in Sec.~\ref{sec:two_properties},
each transaction is then represented by \emph{the dimension indices of the k largest magnitudes of the corresponding image patch}.

This strategy satisfies both requirements given in Sec.~\ref{subsec:FIM_revisited}.
Specifically, due to the sparse nature of CNN activations (sparse property in Sec.\ref{sec:two_properties}), each integer vector transaction calculated as described contains only $k$ items, and $k$ can be set to be small ($20$ in all of our experiments).

Following the work of~\cite{DBLP:conf/iccv/QuackFLG07}, at the end of each transaction, we add a $pos$ ($neg$) item if the corresponding image patch comes from the target category (natural world).
Therefore, each complete transaction has $k+1$ items, consisting of indices of $k$ largest CNN magnitudes plus one class label.
For example, if we set $k$ equals three, given a CNN activation of an image patch from the target category which has $3$ largest magnitudes in its $3rd$, $100th$ and $4096th$ dimension, the corresponding transaction will be $\{3,100,4096,pos\}$.

In practice,  we first sample a large number of patches from images in both the target category and the natural world.
After extracting their CNN activations from \texttt{caffe}, a transaction database $\mathcal{D}$ is created containing a large number of transactions created using the proposed technique above.
Note that the class labels, $pos$ and $neg$, are represented by $4097$
and $4098$ respectively in the transactions.

\subsection{Association rule mining}
\label{subsec:mining}
Given the transaction database $\mathcal{D}$, we use the Aprior algorithm~\cite{DBLP:conf/vldb/AgrawalS94} to discover a set of patterns $\mathcal{P}$ through association rule mining. Each pattern $P \in \mathcal{P}$
must satisfy the following two criteria:
\begin{align}
\label{eq:support}
\supp(P)&>\supp_{\min},\\
\label{eq:confidence}
\conf(P \rightarrow pos)&>\conf_{\min},
\end{align}
where $\supp_{\min}$ and $\conf_{\min}$ are thresholds for the support value and confidence. \vspace{0.2cm}
\newline
\textbf{Representativeness and discriminativeness.} We now demonstrate how association rule mining implicitly satisfies the two requirements of mid-level visual element discovery, \ie, representativeness and discriminativeness.
Specifically, based on Eq.~\ref{eq:support} and Eq.~\ref{eq:confidence}, we are able to rewrite Eq.~\ref{eq:confidence_definition} thus
\begin{equation}
\begin{split}
\supp(P\cup\{pos\})& = \supp(P)\times \conf(P \rightarrow pos)\\
                       & > \supp_{\min}\times \conf_{\min},
\end{split}
\end{equation}
where $\supp(P\cup\{pos\})$ measures the fraction of pattern $P$ found in transactions of the target category among all the transactions.
Therefore, values of $\supp(P)$ and $\conf(P\rightarrow pos)$ above their thresholds ensure that
pattern $P$ is found frequently in the target category, akin to the representativeness requirement.
A high value of $\conf_{\min}$ (Eq.~\ref{eq:confidence}) will also ensure that pattern $P$ is more likely to be found in the target category rather than the natural world, reflecting
the discriminativeness requirement.

\section{Application to image classification}
\label{sec:application}
We now apply our MDPM algorithm to the image classification task.
To discover patterns from a particular class, this class is treated as the target category while all other classes in the dataset
are treated as the natural world.
Note that only training images are used to discover patterns.
\subsection{Retrieving mid-level visual elements}
Given the pattern set $\mathcal{P}$ discovered in Sec.~\ref{sec:DFIM},
finding mid-level visual elements is straightforward.  A mid-level visual element $V$ contains the image patches sharing the same pattern $P$, which can be retrieved efficiently through an inverted index. This process gives rise to a set of mid-level visual elements $\mathcal{V}$ (\ie $V \in \mathcal{V}$).

We provide a visualization of some of the visual elements discovered by the MDPM in Fig.~\ref{fig:Name_that_scene}. It is clear that image patches in each visual element are visually similar and depict similar semantic concepts.
An interesting observation is that visual elements discovered by the MDPM are invariant to horizontal flipping.

\subsection{Ensemble merging and training detectors}
\label{subsec:training_detectors}
We note that patches belonging to different elements may overlap or describe the same visual concept. To remove this redundancy, we propose to merge elements in an iterative procedure while training element detectors.

Algorithm~\ref{alg:merging} summarizes the proposed ensemble merging procedure.
At each iteration, we greedily merge overlapping mid-level elements and train the corresponding detector through the \texttt{MergingTrain} function in Algorithm~\ref{alg:merging}.
In the \texttt{MergingTrain} function, we begin by selecting the element covering the maximum number of training images, followed by training a Linear Discriminant Analysis (LDA) detector~\cite{DBLP:conf/eccv/HariharanMR12}.
We then incrementally revise this detector. At each step, we run the current detector on the patches of all the remaining visual elements, and retrain it by augmenting the positive training set with positive detections. We repeat this iterative procedure until no more elements can be added into the positive training set.
The idea behind this process is using detection score as a similarity metric, much inspired by Exemplar SVM~\cite{DBLP:conf/iccv/MalisiewiczGE11, DBLP:journals/tog/ShrivastavaMGE11}.

The final output of the ensemble merging step is a clean set of visual elements and their corresponding detectors.
\begin{algorithm}[t]
\label{alg:merging}

\SetKwFunction{Grow}{Grow}
\SetKwFunction{MergingTrain}{MergingTrain}
\SetKwFunction{Coverage}{Coverage}
\SetKwFunction{Score}{Score}
\caption{Ensemble Merging Pseudocode}
\KwIn{A set of partially redundant visual elements $\mathcal{V}$}
\KwOut{A set of clean mid-level visual elements $\mathcal{V}'$ and corresponding element detectors $D$}
Initialize $\mathcal{V}' \leftarrow \emptyset$, $D_c \leftarrow \emptyset$\;
\While{$\mathcal{V} \neq \emptyset$}{
$[\mathcal{V}_t,d] \leftarrow \MergingTrain(\mathcal{V})$\;
$\mathcal{V} \leftarrow \mathcal{V} \setminus \mathcal{V}_t$\;
$\mathcal{V}' \leftarrow \mathcal{V}' \cup \{ \mathop{\cup}\limits_{V \in \mathcal{V}_t} V\}$\;
$D \leftarrow D \cup \{d\}$\;
}
\KwRet $\mathcal{V}'$, $D$\;
\BlankLine
\SetKwProg{myFunc}{Function}{}{}
\myFunc{\MergingTrain{$\mathcal{V}$}}{
  Select $V^{\ast} \in \mathcal{V}$ which covers the maximum number of training images\;
  Initialize $\mathcal{V}_t \leftarrow \{V^{\ast}\}$, $\mathcal{S} \leftarrow \emptyset$ \;

	\Repeat{$\mathcal{S} = \emptyset$}{
  $\mathcal{V}_t \leftarrow \mathcal{V}_t \cup \mathcal{S}$\;
  Train LDA detector $d$ using $\mathcal{V}_t$\;
  $\mathcal{S} \leftarrow \{V \in \mathcal{V}\setminus\mathcal{V}_t | \Score(V,d)>Th$\} where
  $\Score(V,d) = \frac{1}{|V|}\sum_{x \in V} d^Tx$ ($Th$ is a pre-defined threshold, $x$ is a CNN activation of an image patch in $V$)\;
	}
  \KwRet $\mathcal{V}_t$, $d$\;
}

\end{algorithm}
\subsection{Selecting and encoding}
We can now use the learned element detectors to encode a new image.
There is a computational cost, however, associated with applying each successive learned element, and particular elements may be more informative when applied to particular tasks.  We thus now seek to identify those elements of most value to the task at hand.

In practice, we rank all of the elements in a class based on the number of training images that they cover.
We then select the detectors corresponding to the elements which cover the maximum number of training images, akin to ``maximizing coverage'' in~\cite{DBLP:conf/nips/DoerschGE13}.
This process is then repeated such that the same number of detectors are selected from each class and stacked together.

To generate a feature representation for a new image, we evaluate all of the selected detectors at three scales.
For each scale, we take the max score per detector per region encoded in a $2$-level ($1\times1$ and $2\times2$) spatial pyramid.
The final feature vector is the outcome of max pooling on the features from all three scales.

\section{Experiments}
We test our algorithm on two image classification tasks, scene classification and generic object classification.

\textbf{Implementation details.}
For each image, we resize its smaller dimension to $256$ while maintaining its aspect ratio, then we sample $128\times128$ patches with a stride of $32$ pixels, and calculate the CNN activations using \texttt{caffe}. Because the number of patches sampled varies in different datasets, two parameters $\supp_{\min}$ and $\conf_{\min}$ in the association rule mining (Sec.~\ref{subsec:mining}) are set according to each dataset with the goal that at least $100$ patterns are discovered for each category.
The merging threshold $Th$ in Algorithm~\ref{alg:merging} (Sec.~\ref{subsec:training_detectors}) is set as $150$.
For training classifiers, we use the Liblinear toolbox~\cite{DBLP:journals/jmlr/FanCHWL08}
with $5$-fold cross validation.

\subsection{Scene classification}
The MIT Indoor dataset~\cite{DBLP:conf/cvpr/QuattoniT09} contains 67 classes of indoors scenes.
Verified by recent works on mid-level visual element discovery, indoor scenes are better characterized by the unique objects that they contain
(\eg, computers are more likely to be found in computer room rather than laundry).
We follow the standard partition of~\cite{DBLP:conf/cvpr/QuattoniT09}, \ie, approximately 80 training and 20 test images per class.
$\supp_{\min}$ and $\conf_{\min}$ are set as $0.01\%$ and $30\%$ respectively.

\vspace{2mm}
\noindent\textbf{Comparison to methods using mid-level visual elements.} We first compare our approach against recent works on mid-level visual element discovery (See Table~\ref{tab:discrminativePatch}). Using only $20$ visual elements per class, our approach yields an accuracy of $68.24$\%.
Increasing the number of visual elements to $50$ makes our performance increases to $69.69$\%, outperforming all previous mid-level visual element discovery methods by a sizeable margin.
As shown in Table~\ref{tab:discrminativePatch},
compared with the work of Doersch \etal~\cite{DBLP:conf/nips/DoerschGE13}
which achieved best performance among previous mid-level visual elements algorithms, our approach uses an order of magnitude fewer elements than~\cite{DBLP:conf/nips/DoerschGE13} ($20$ \emph{vs.} $200$) while outperforming it by over $4$ percent in accuracy.
Also, our approach surpasses a very recent work  RFDC~\cite{DBLP:conf/eccv/BossardGG14} by over $15\%$ in the same setting ($50$ elements per class).
Thanks to the fact that CNN activations from \texttt{caffe} are invariant to horizontal flipping~\cite{DBLP:conf/eccv/GongWGL14}, we avoid adding right-left flipped images (\cf~\cite{DBLP:conf/cvpr/JunejaVJZ13,DBLP:conf/nips/DoerschGE13}).

\vspace{2mm}
\noindent\textbf{Ablation study}. As we are the first to use CNN activation for mid-level visual elements discovery, a natural question is that what is the performance of previous
works if CNN activation is adopted? To answer this question, we implement
two baselines using CNN activations as image patch representations. ``LDA-Retrained'' initially trains Exemplar LDA using a sampled patch and then re-trains the detector $10$ times by adding top-$10$ positive detections as positive training samples at each iteration. This is quite similar to the ``Expansion'' step of~\cite{DBLP:conf/cvpr/JunejaVJZ13}. Another baseline ``LDA-KNN'' retrieves 5-nearest neighbors of an image patch and trains a LDA detector using the retrieved patches (including itself) as positive training data. For both baselines, discriminative detectors are selected based on the Entropy-Rank Curves proposed in~\cite{DBLP:conf/cvpr/JunejaVJZ13}.
As shown in Table~\ref{tab:discrminativePatch}, we report much better results than both baselines in the same setting, which verifies that the proposed MDPM is an essential step for achieving good performance when using CNN activations for mid-level visual element discovery.
\begin{table}
\begin{center}
\begin{tabular}{|l|c|c|c|}
\hline
Method & $\#$elements & Acc(\%)\\
\hline
D-patch~\cite{DBLP:conf/eccv/SinghGE12} & $210$ & $38.10$\\
BoP~\cite{DBLP:conf/cvpr/JunejaVJZ13} & $50$ & $46.10$\\
miSVM~\cite{DBLP:conf/cvpr/LiWT13} & $20$ & $46.40$\\
MMDL~\cite{DBLP:conf/icml/WangWBLT13} & $11$ & $50.15$\\
D-Parts~\cite{DBLP:conf/iccv/SunP13} & $73$ & $51.40$\\
RFDC~\cite{DBLP:conf/eccv/BossardGG14} & $50$ & $54.40$\\
DMS~\cite{DBLP:conf/nips/DoerschGE13} & $200$ & $64.03$\\
\hline
LDA-Retrained & $20$ & $58.78$\\
LDA-Retrained & $50$ & $62.30$\\
LDA-KNN & $20$ & $59.14$\\
LDA-KNN & $20$ & $63.93$\\
\hline
Ours & $20$ & $68.24$\\
Ours & $50$ & \boldsymbol{$69.69$}\\
\hline
\end{tabular}
\end{center}
\caption{Classification results of mid-level visual element discovery algorithms on the MIT Indoor dataset.}
\label{tab:discrminativePatch}
\end{table}

\begin{table}[t]
\begin{center}
\begin{tabular}{|l|c|l|}
\hline
Method & Acc(\%)  & Comments\\
\hline
CNN-G & $57.74$ & CNN for whole image\\
CNN-Avg & $65.77$ & average pooling\\
CNN-Max & $65.15$ & max pooling\\
\hline
CNN-SVM~\cite{6910029}  & $58.40$ & \texttt{OverFeat} toolbox\\
VLAD level2~\cite{DBLP:conf/eccv/GongWGL14} & $65.52$ & VLAD encoding\\
VLAD level3~\cite{DBLP:conf/eccv/GongWGL14} & $62.24$ & VLAD encoding\\
VLAD level1\&2~\cite{DBLP:conf/eccv/GongWGL14} & $66.64$ & concatenation\\
MOP-CNN~\cite{DBLP:conf/eccv/GongWGL14} & $68.88$ & concatenation\\
CNN-jittered~\cite{DBLP:journals/corr/AzizpourRSMC14} & $65.50$ & jittered CNN \\
CNN-finetune~\cite{DBLP:journals/corr/AzizpourRSMC14} & $66.30$ & fine-tuned CNN \\
Places-CNN~\cite{DBLP:conf/nips/ZhouLXTO14} & $68.24$ & Places dataset used\\
SCFVC~\cite{DBLP:conf/nips/LiuSWHW14} & $68.20$ & new Fisher encoding\\
CL-45C~\cite{DBLP:conf/cvpr/LiuSH15} & $68.80$ & cross-layer pooling\\
\hline
Ours & $69.69$ & MDPM ($50$ element)\\
Ours+CNN-G & \boldsymbol{$70.46$} & concatenation\\
\hline
\end{tabular}
\end{center}
\caption{Classification results of methods using CNN activations on the MIT Indoor dataset.}\vspace{-3mm}
\label{tab:VLAD_comparison}
\end{table}

\begin{figure*}[t]
\vspace{-0.0cm}
\begin{center}
\begin{tabular}{@{}c@{}c@{}c@{}c}
\includegraphics[width=0.25\linewidth]{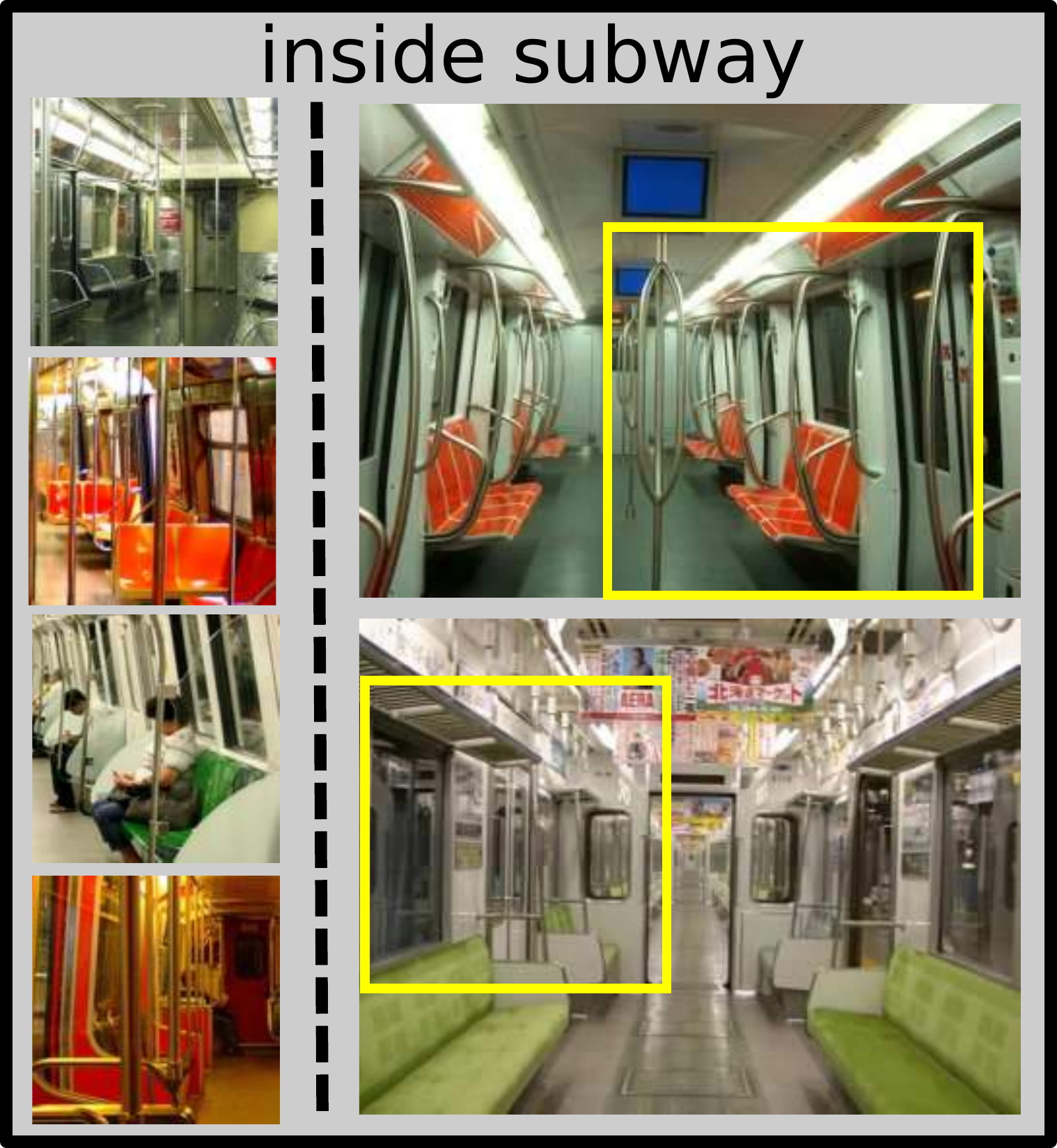} \ &
\includegraphics[width=0.25\linewidth]{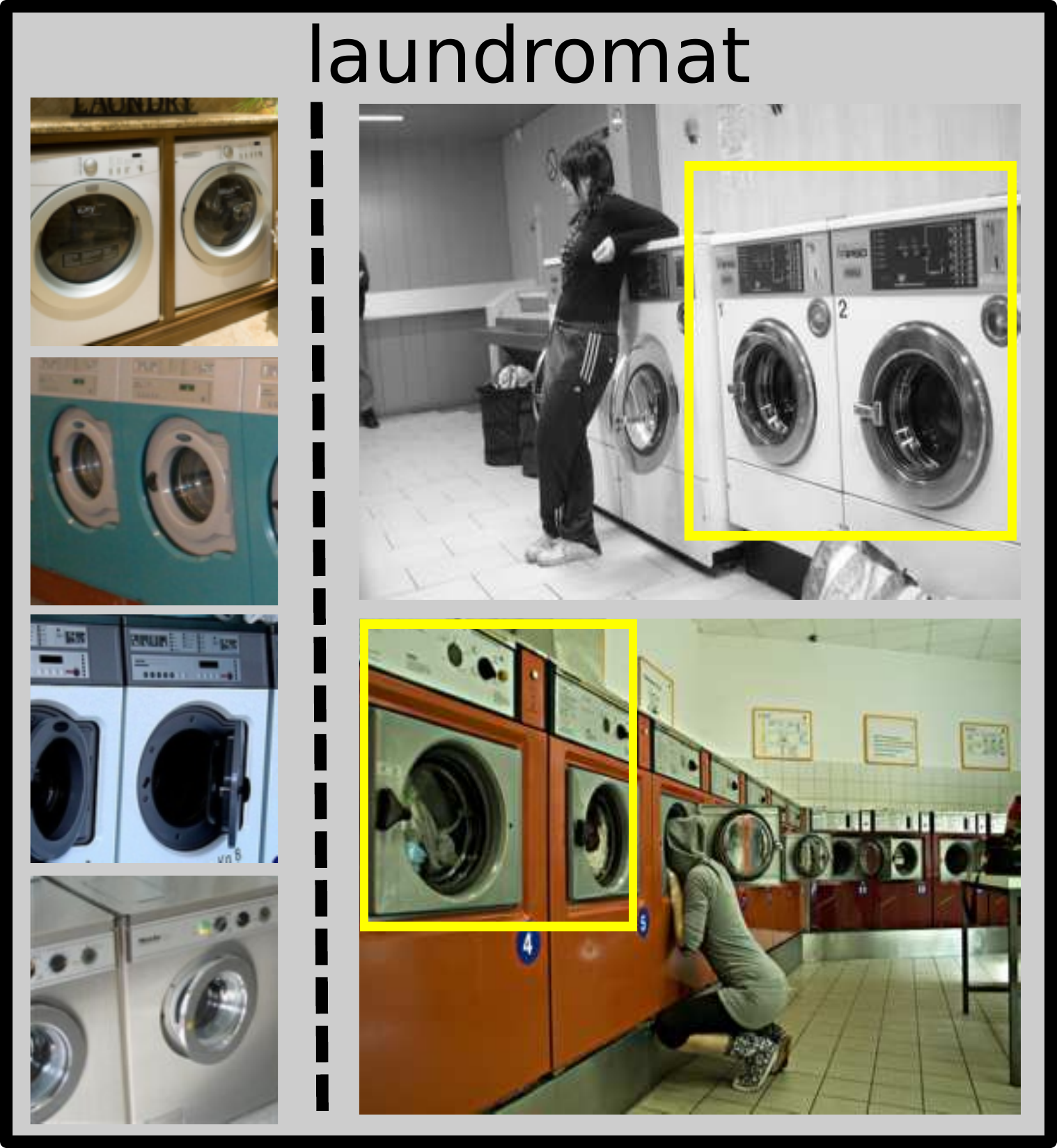} \ &
\includegraphics[width=0.25\linewidth]{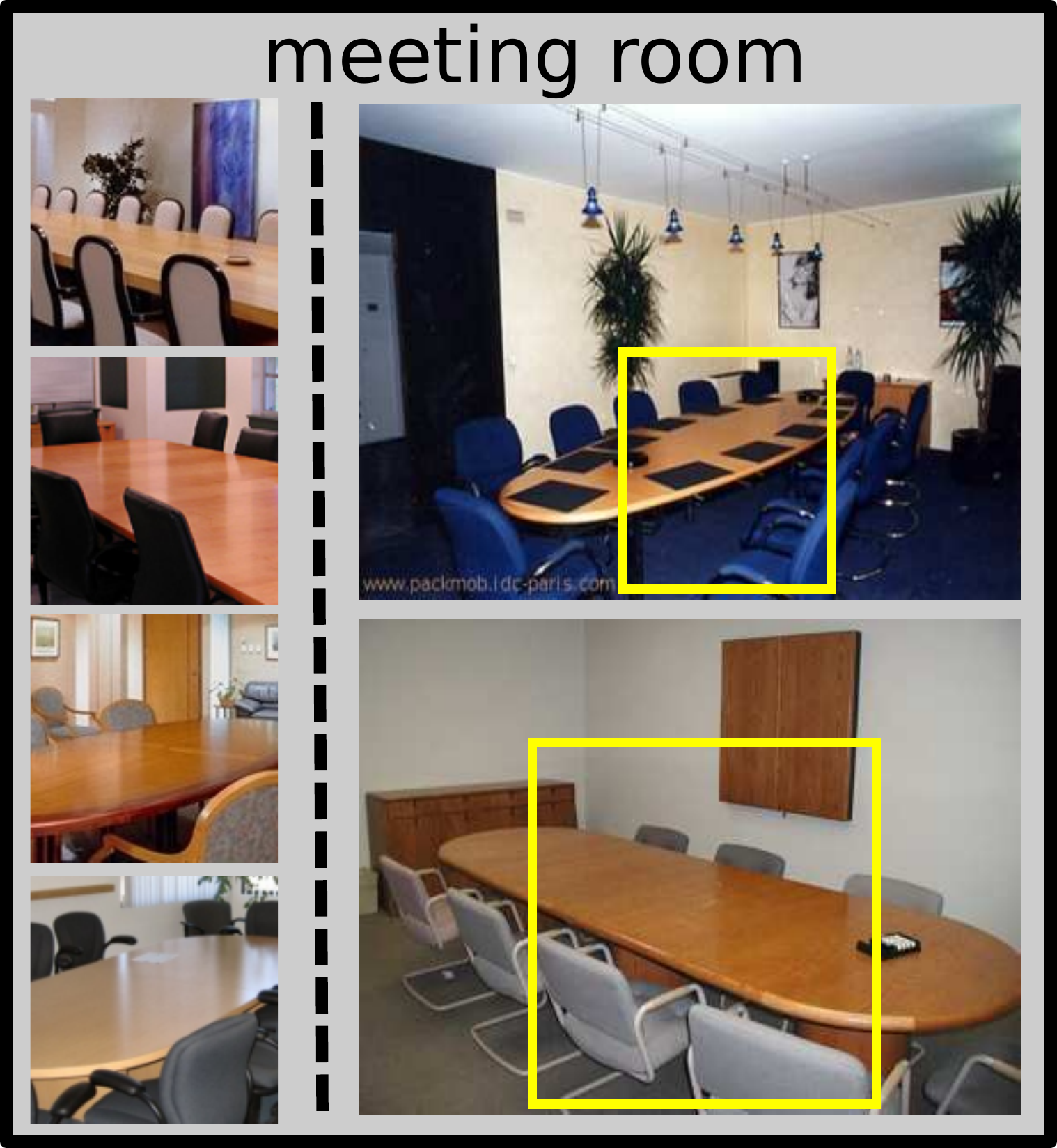} \ &
\includegraphics[width=0.25\linewidth]{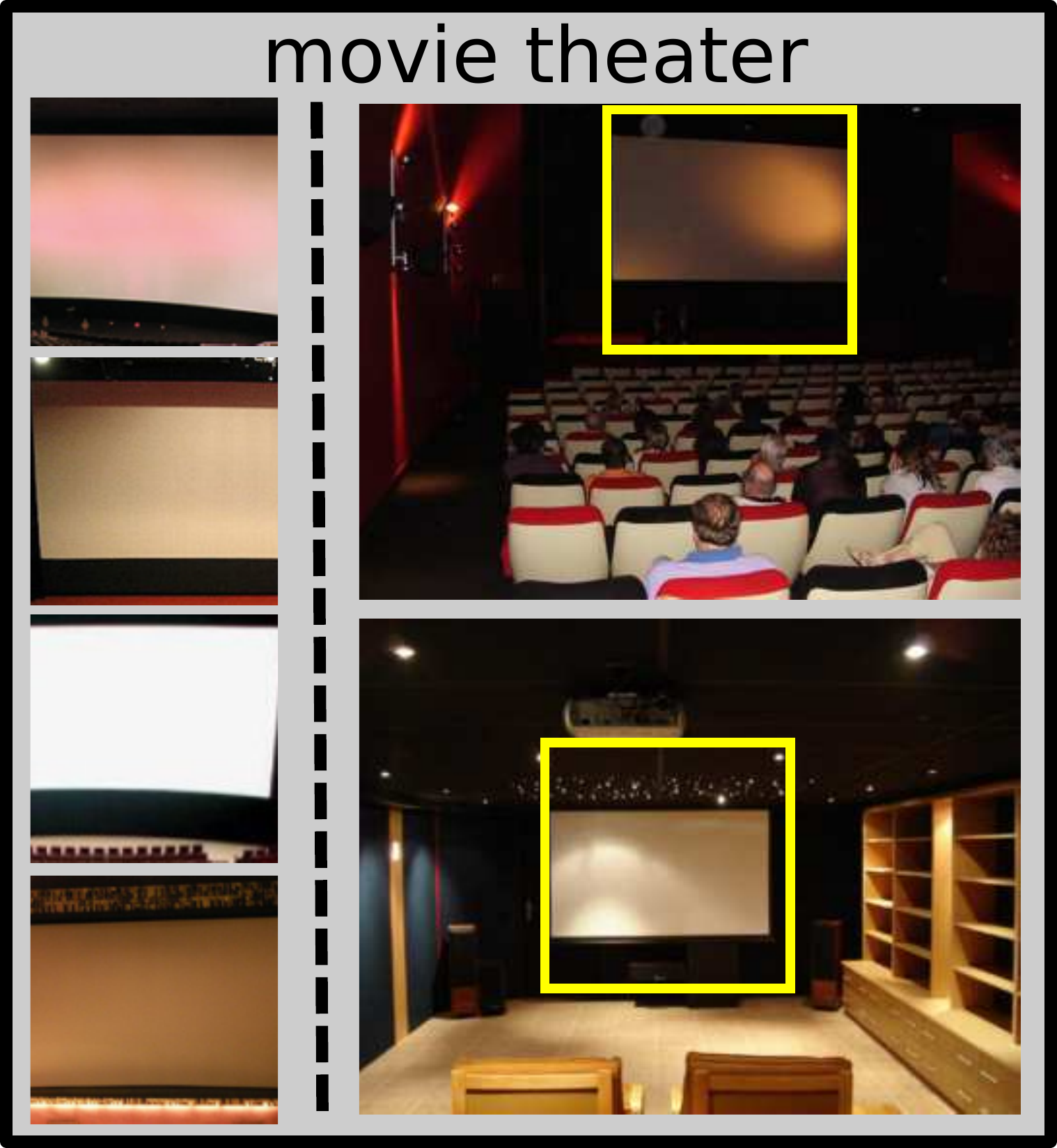} \ \\
\includegraphics[width=0.25\linewidth]{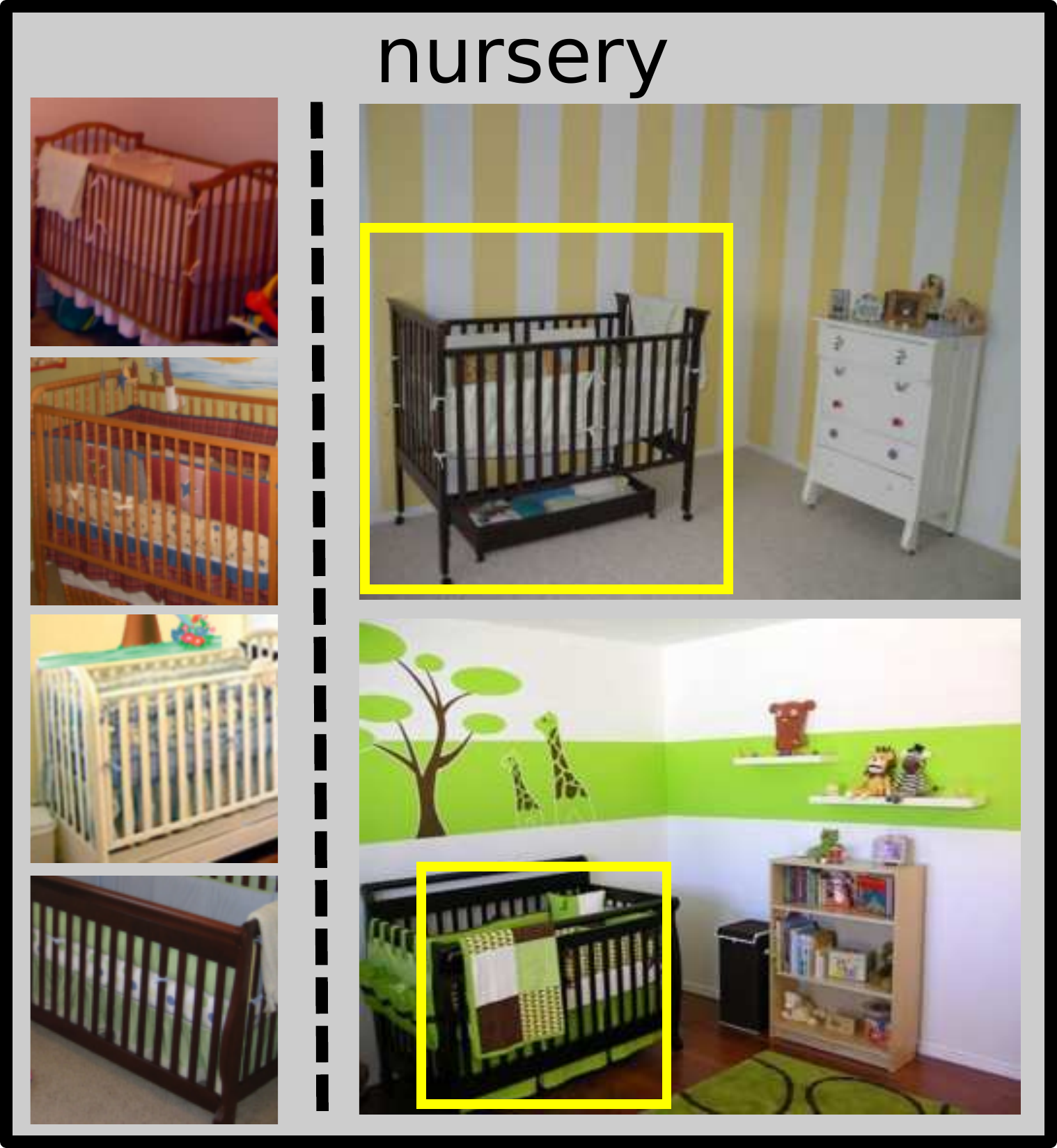} \ &
\includegraphics[width=0.25\linewidth]{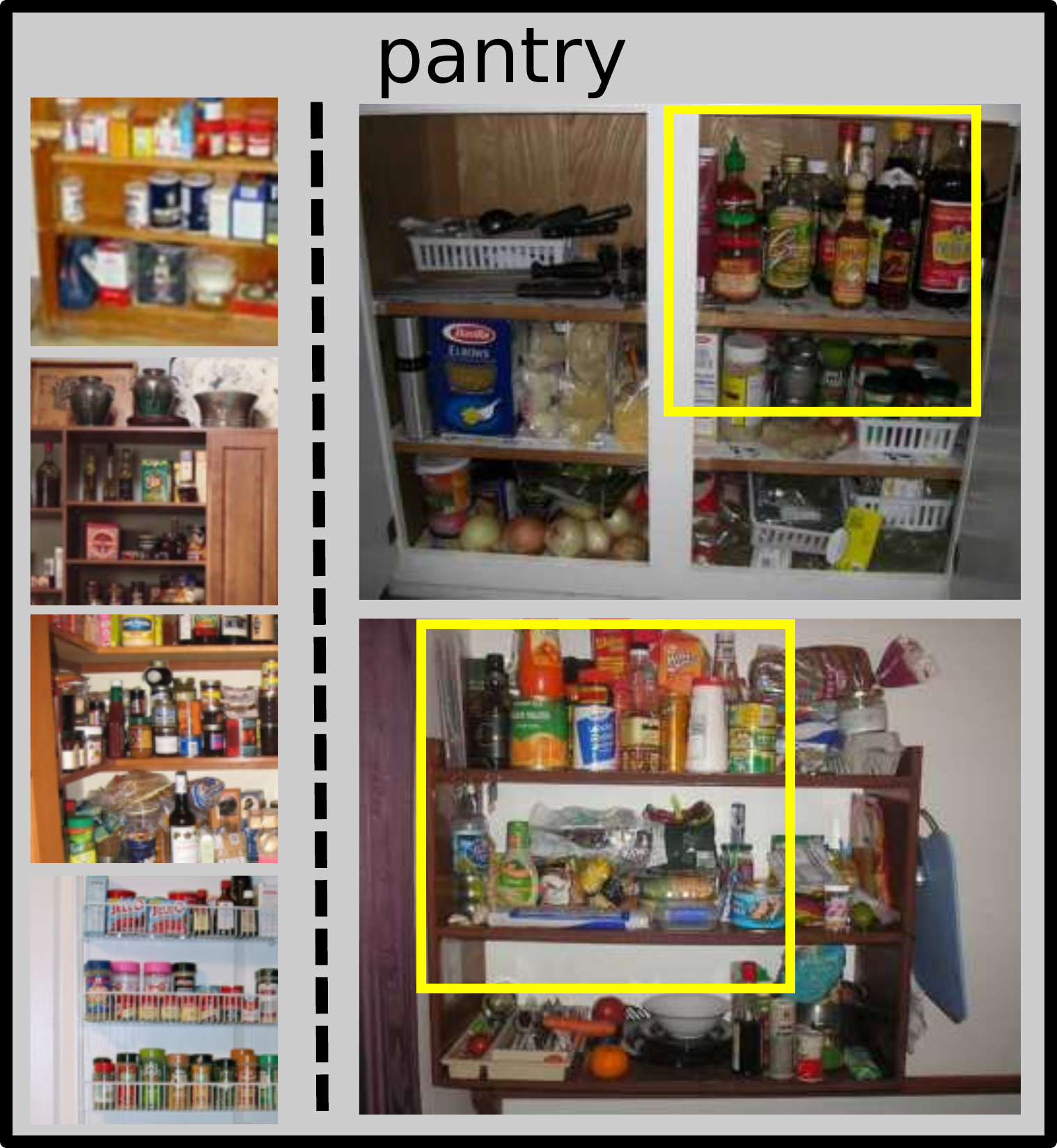} \ &
\includegraphics[width=0.25\linewidth]{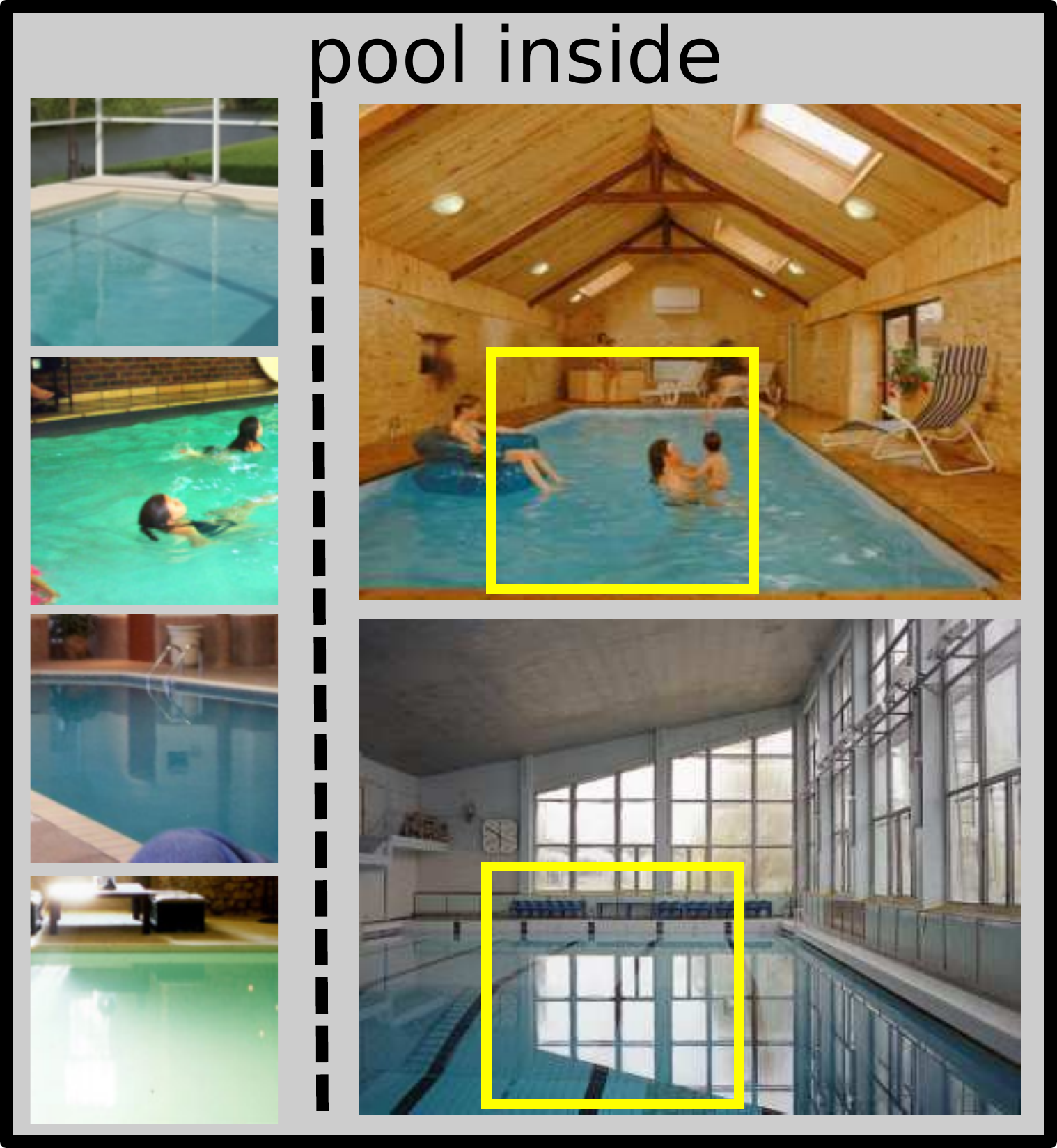} \ &
\includegraphics[width=0.25\linewidth]{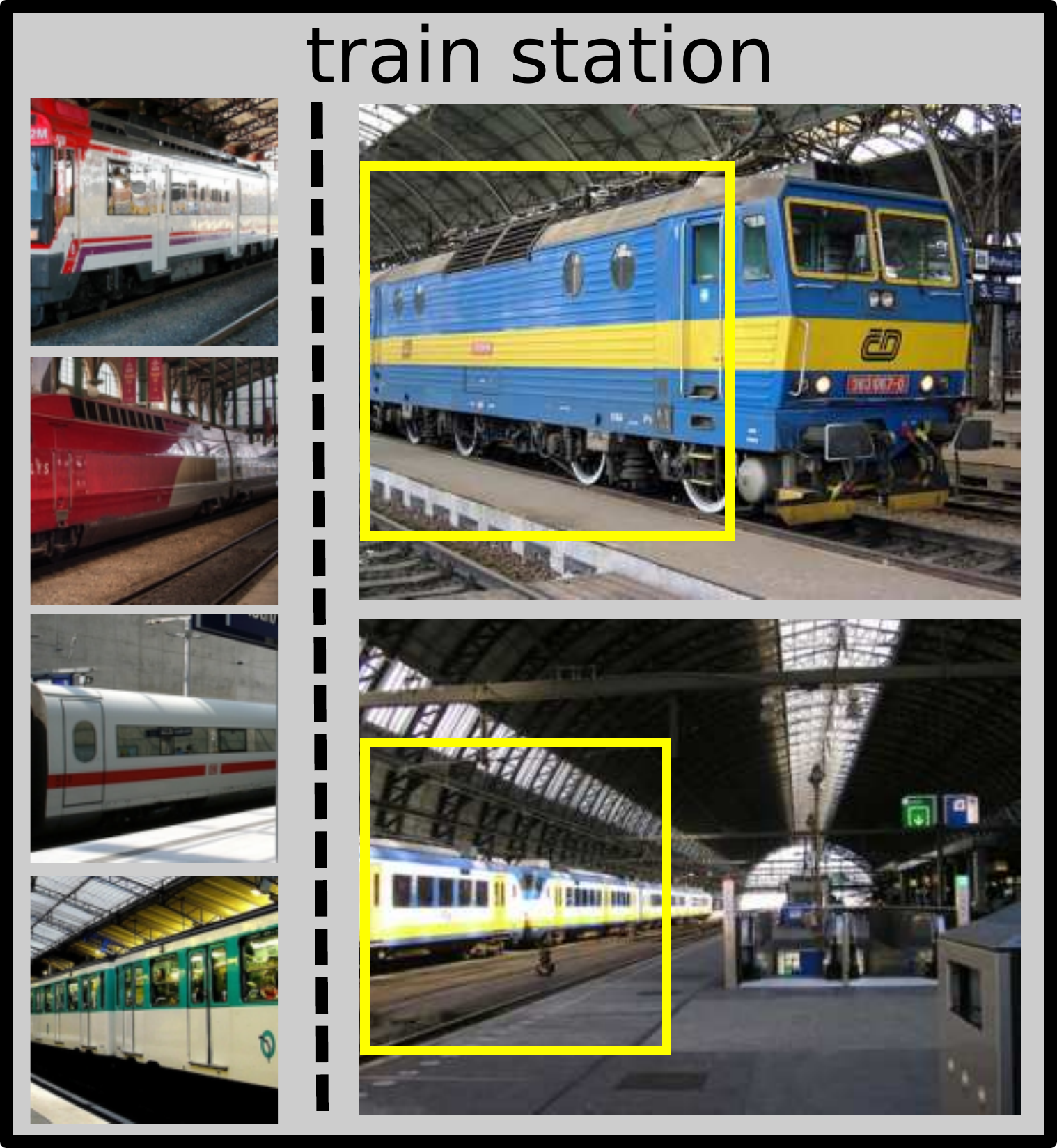} \ \\
\end{tabular}
\end{center}
\caption{Discovered mid-level visual elements and their corresponding detections on test images on the MIT Indoor dataset.}
\label{fig:fire_MIT}
\end{figure*}

\vspace{2mm}
\noindent\textbf{Comparison to methods using CNN activation}.
In Table~\ref{tab:VLAD_comparison}, we compare our approach to others in which CNN activation is involved.
Our baseline method, using \texttt{fc}${6}$ CNN activations extracted from the whole image, gives an accuracy of $57.74\%$.
Our approach (based on the MDPM algorithm) achieves $69.69\%$ accuracy, which   is a significant improvement over all the baselines.
Our work is most closely related to MOP-CNN~\cite{DBLP:conf/eccv/GongWGL14} and SCFVC~\cite{DBLP:conf/cvpr/LiuSH15} in the sense that all these works rely on off-the-shelf CNN activations of image patches.
To encode these local CNN activations, MOP-CNN ~\cite{DBLP:conf/eccv/GongWGL14} rely on the classical VLAD encoding, whereas
SCFVC~\cite{DBLP:conf/cvpr/LiuSH15} is a new Fisher vector encoding strategy for encoding high-dimensional local features.
Our encoding method, which is based on the discovered visual elements, not only outperforms MOP-CNN\cite{DBLP:conf/eccv/GongWGL14} on $128\times128$ and $64\times64$ patches by a noticeable margin ($69.69\%$ \emph{vs.} $65.52\%$ and $69.69\%$ \emph{vs.} $62.24\%$), also slightly bypasses that of SCFVC ($69.69\%$ \emph{vs.} $68.20\%$).

Fine-tuning has been shown to be beneficial when transferring pre-trained CNN models on the ImageNet to another dataset~\cite{DBLP:conf/cvpr/GirshickDDM14,DBLP:conf/cvpr/OquabBL14,
DBLP:conf/eccv/Agrawal14}. Jittered CNN features (\eg, crops, flips) extracted from the fine-tuned network of Azizpour \etal~\cite{DBLP:journals/corr/AzizpourRSMC14} offer $66.3$\% accuracy, which is still below ours.

After concatenating with CNN activations of the whole image (both normalized to unit norm), our performance increases to $70.46\%$, outperforming all previous works using CNN on this dataset.

\vspace{2mm}
\noindent\textbf{Computational complexity.} Given pre-computed CNN activation from about 0.2 million patches, the baseline method ``LDA-Retrained'' takes about 9 hours to find visual elements in a class.
However, our approach only takes about 3 minutes (writing transaction file and association rule mining) to discover representative and discriminative rules.

\vspace{2mm}
\noindent\textbf{Visualization.} We visualize some visual elements discovered and their firings on test images in Fig.~\ref{fig:fire_MIT}.
It is intuitive that the discovered mid-level visual elements capture the patterns which are often repeated within a scene category.
Some of the mid-level visual elements  refer to frequently occurring objects configurations, \eg, the configuration between table and chair in the \emph{meeting room} category.
Some instead capture a particular object in the scene, such as the baby cot
in the \emph{nursery} and screen in the \emph{movie theater}.
\subsection{Object classification}
\begin{figure*}[t]
\vspace{-0.0cm}
\begin{center}
\begin{tabular}{@{}c@{}c@{}c@{}c}
\includegraphics[width=0.25\linewidth]{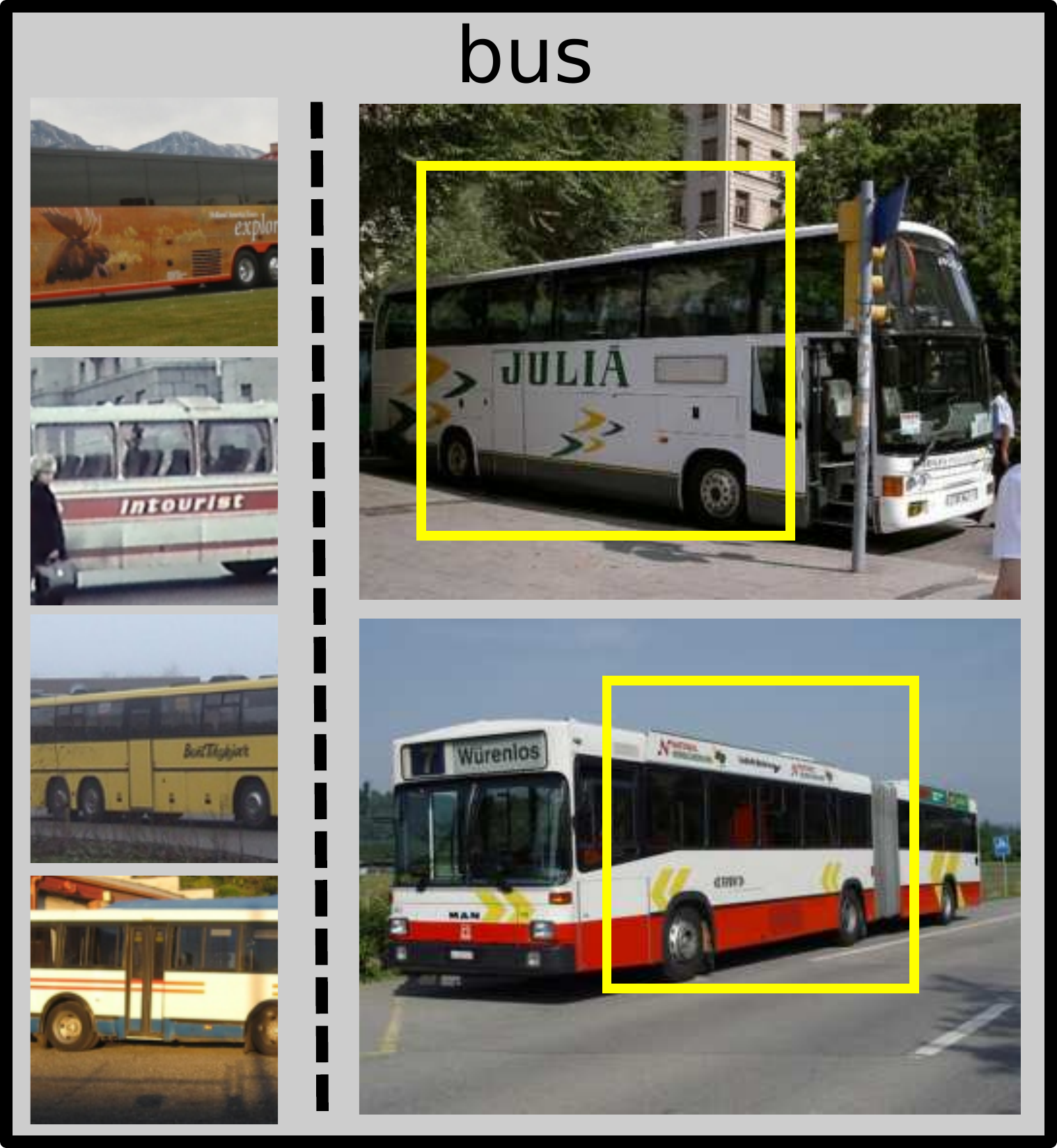} \ &
\includegraphics[width=0.25\linewidth]{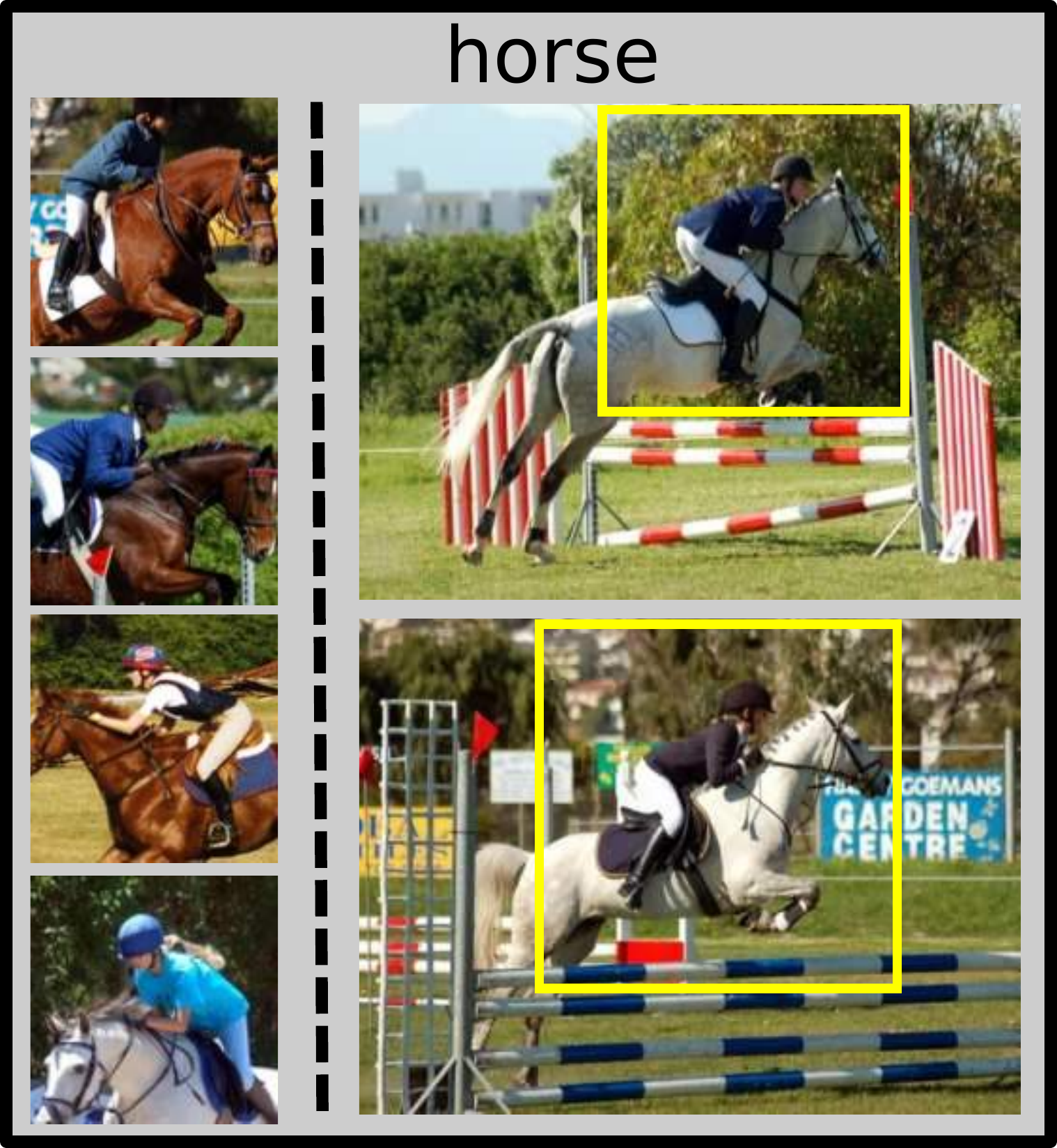} \ &
\includegraphics[width=0.25\linewidth]{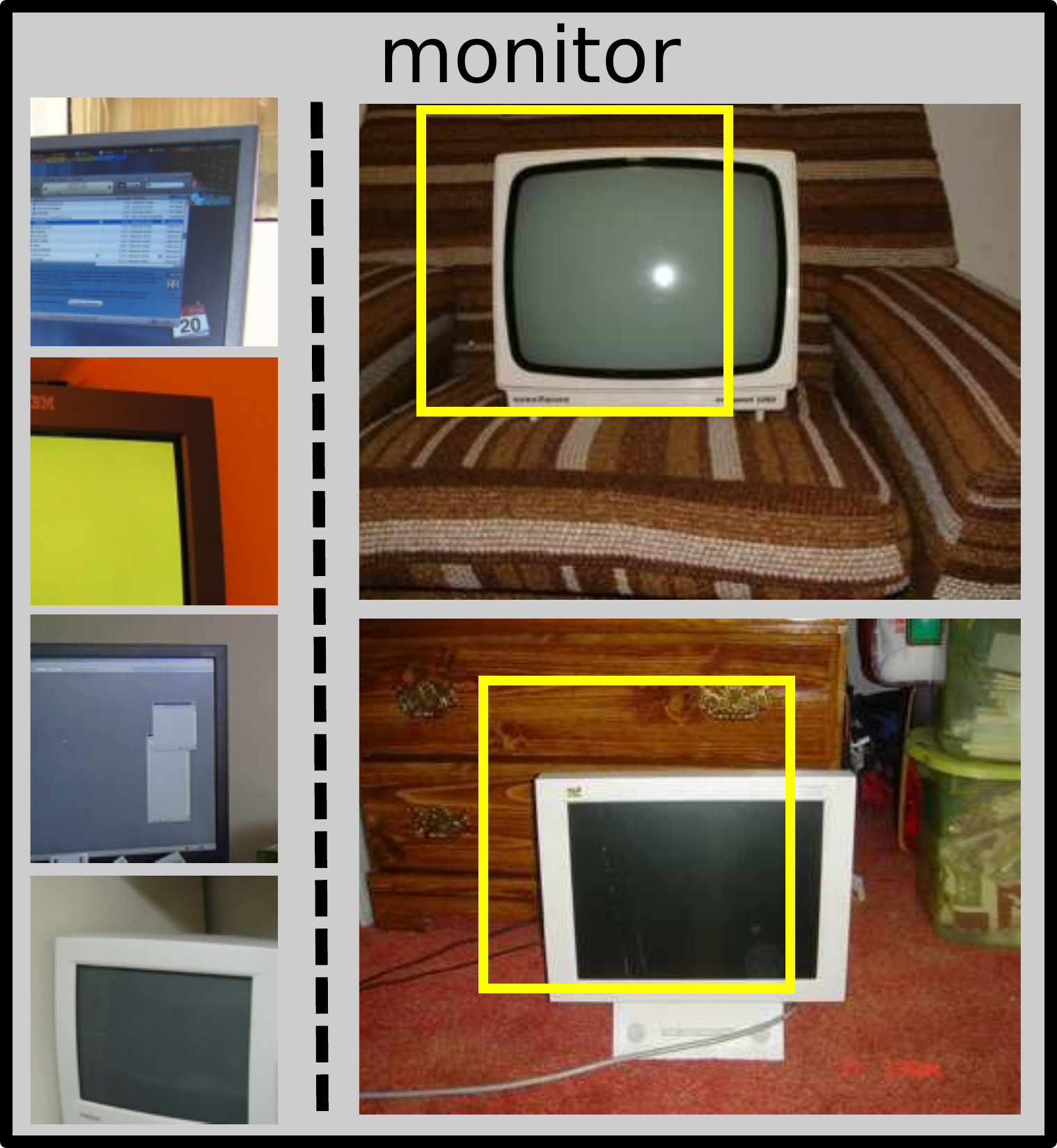} \ &
\includegraphics[width=0.25\linewidth]{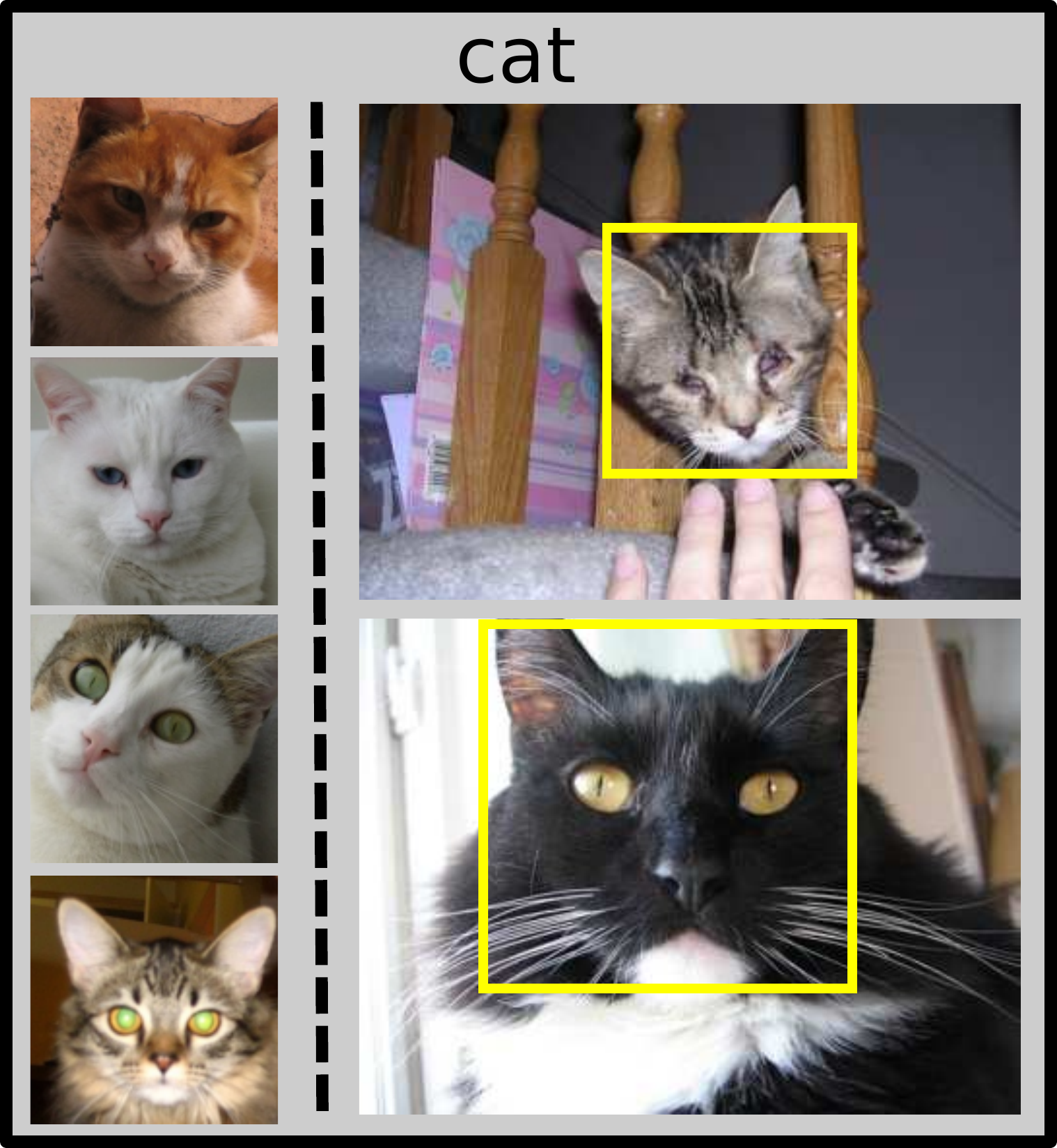} \ \\
\includegraphics[width=0.25\linewidth]{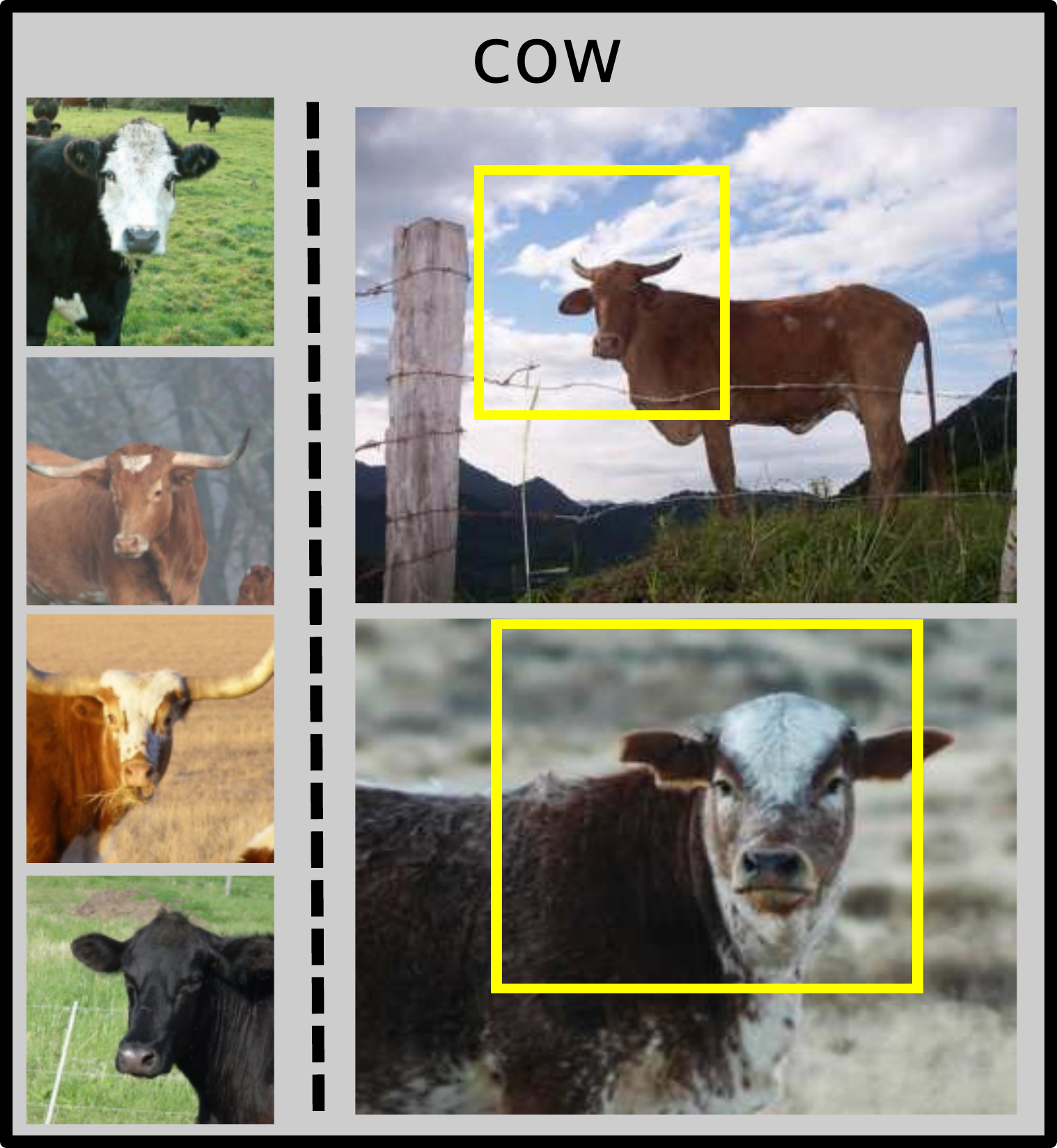} \ &
\includegraphics[width=0.25\linewidth]{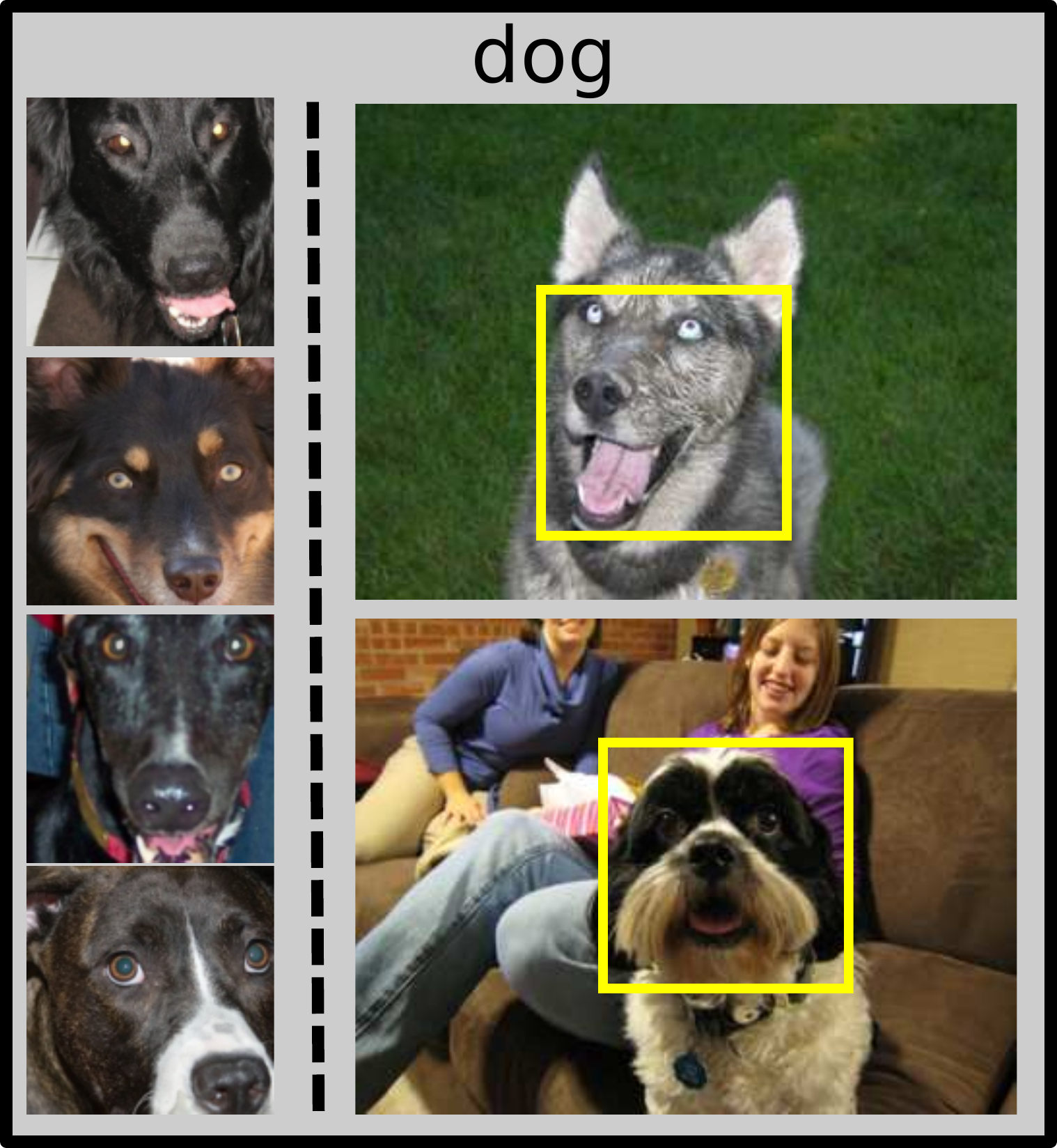} \ &
\includegraphics[width=0.25\linewidth]{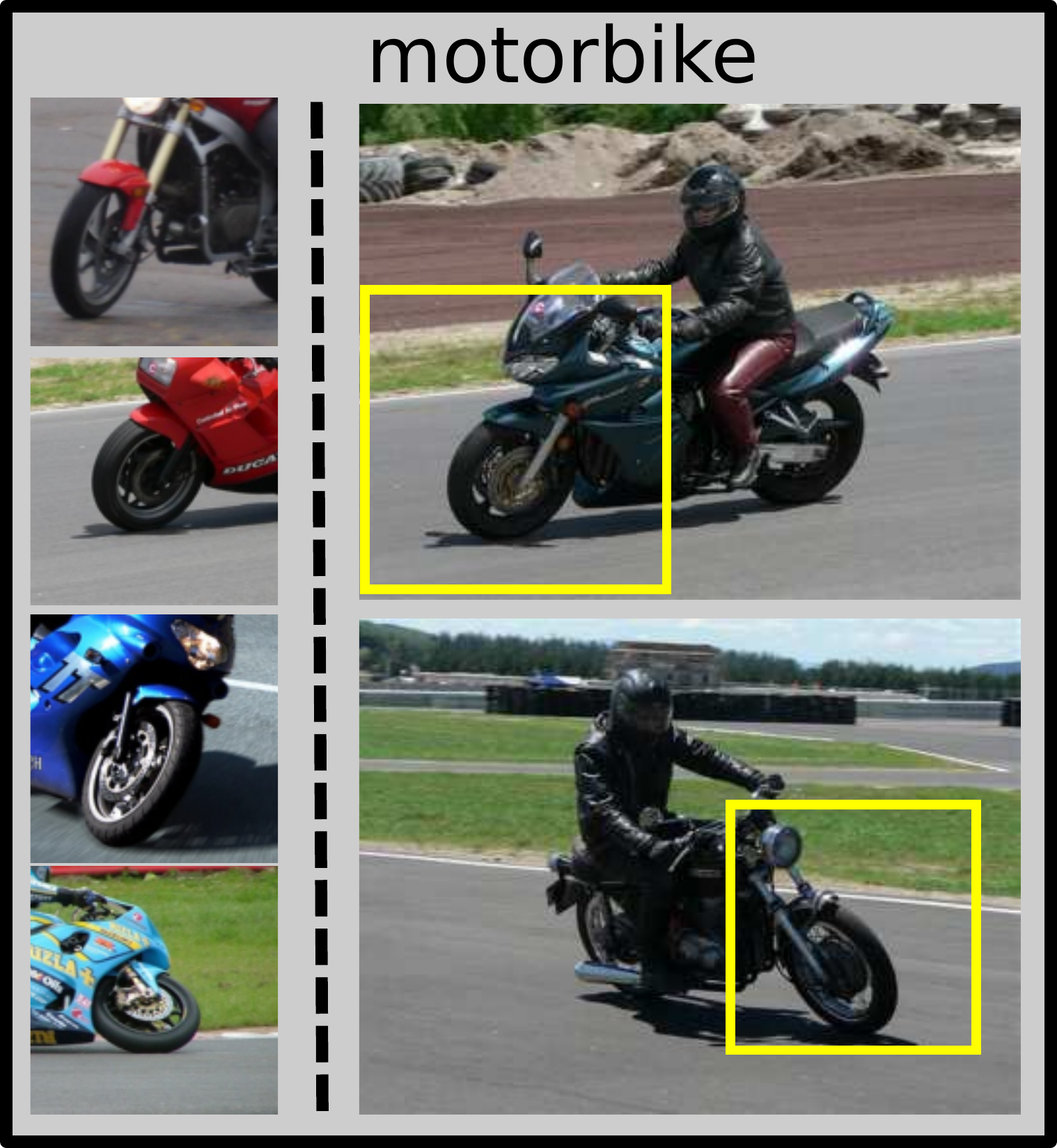} \ &
\includegraphics[width=0.25\linewidth]{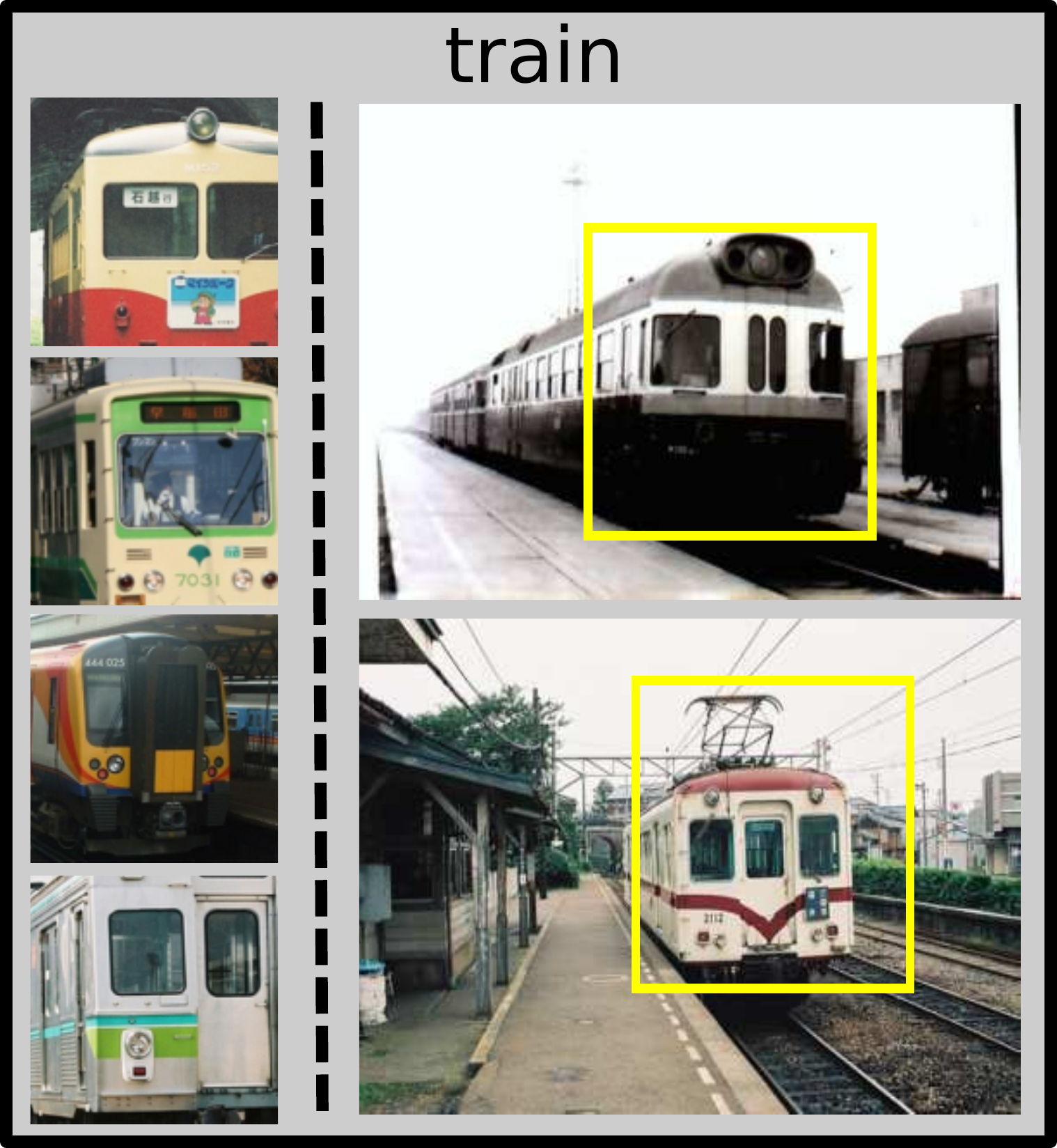} \ \\

\end{tabular}
\end{center}
\caption{Discovered mid-level visual elements and their corresponding detections on test images on the PASCAL VOC 2007 dataset.}
\label{fig:fire_VOC}
\end{figure*}
\begin{table}[t]
\begin{center}
\begin{tabular}{|l|c|l|}
\hline
Method & mAP(\%) & Comments\\
\hline
CNN-G & $67.3$ & CNN for whole image\\
CNN-Avg & $70.9$ & average pooling\\
CNN-Max & $71.1$ & max pooling\\
\hline
CNN-SVM~\cite{6910029} & $73.9$ & \texttt{OverFeat} toolbox\\
PRE-1000C~\cite{DBLP:conf/cvpr/OquabBL14} & $77.7$ & bounding box used\\
CNN-jittered~\cite{DBLP:journals/corr/AzizpourRSMC14} & $72.2$ & jittered CNN \\
SCFVC~\cite{DBLP:conf/nips/LiuSWHW14} & $76.9$ & new Fisher encoding\\
CL-45C~\cite{DBLP:conf/cvpr/LiuSH15} & $75.0$ & cross-layer pooling\\
\hline
Ours  & $75.2$ & MDPM (50 elements)\\
Ours+CNN-G & \boldsymbol{$77.9$} & concatenation\\
\hline
\end{tabular}
\end{center}
\caption{Classification results of methods using CNN activations on the PASCAL VOC 2007 dataset.}\vspace{-3mm}
\label{tab:VOC2007_result}
\end{table}
The Pascal VOC 2007 dataset~\cite{DBLP:journals/ijcv/EveringhamGWWZ10} contains $9,963$ images from 20 object classes.
For this dataset, training and validation sets are utilized to discover patterns and training final classifiers.
The parameters $\supp_{\min}$ and $\conf_{\min}$ are set as $0.01\%$ and $60\%$ respectively.

\vspace{2mm}
\noindent\textbf{Comparison to methods using CNN activation.} Table~\ref{tab:VOC2007_result} reports our results along with those of other recent methods based on CNN activation.
On this dataset, when using $50$ visual elements per class, the proposed method achieves $75.2\%$ mean average precision (mAP), significantly outperforming the baseline that using CNN activations as a global feature ($67.3\%$), as well as its average pooling and max pooling counterparts.
Compared with state-of-the-arts,  %
Oquab~\etal~\cite{DBLP:conf/cvpr/OquabBL14} fine-tune the pre-trained network on the ImageNet, however, it relies on bounding box annotation which makes the task easier, so it is not surprising that it outperforms ours which does not use bounding box annotation.
The best result on PASCAL VOC 2007 ($77.9\%$) is achieved when the proposed MDPM feature and global CNN activation are concatenated, marginally outperforming fine-tuning with bounding box annotation ~\cite{DBLP:conf/cvpr/OquabBL14}.  This is despite the fact that the bounding box annotations constitute extra information which is time-consuming to gather.

\vspace{2mm}
\noindent\textbf{Visualization.} We visualize some visual elements discovered and their firings on
the test images of the VOC 2007 dataset in Fig.~\ref{fig:fire_VOC}.
It is clear that the discovered mid-level visual elements capture some discriminative parts of object (\eg, dog faces).
It is worth noting here that ``parts'' have been shown to be extremely important for state-of-the-art object recognition, such as Deformable Part Models~\cite{DBLP:journals/pami/FelzenszwalbGMR10} and Poselets~\cite{DBLP:conf/iccv/BourdevM09}.
\section{Conclusion}
We have addressed mid-level visual element discovery from the perspective of pattern mining.
In the process we have shown not only that is it profitable to apply pattern mining technique to mid-level visual element discovery, but also that, from the right perspective, CNN activations are particularly well suited to the task.
This is significant because CNNs underpin many current state-of-the-art methods in vision, and pattern mining underpins significant elements of the state-of-the-art in Big Data processing.

{\small
\bibliographystyle{ieee-cs}
\bibliography{egbib}
}

\end{document}